\documentclass[twoside]{article}

\usepackage[accepted]{aistats2025}
\usepackage{cite}
\usepackage{hyperref}
\usepackage{float}
\usepackage{natbib}
\usepackage{amsmath,amssymb,amsfonts}
\usepackage{algorithmic}
\usepackage{graphicx}
\usepackage{textcomp}
\usepackage{xcolor}
\usepackage{hyperref}
\usepackage{url}
\usepackage{amssymb}
\usepackage{amsmath}
\usepackage{dsfont}
\usepackage{tcolorbox}
\usepackage{graphicx} 
\usepackage{multirow}
\usepackage{bbding}
\usepackage{pifont}
\usepackage{algorithm}
\usepackage{algorithmic}
\usepackage{graphicx}
\usepackage{subcaption}
\usepackage{booktabs}
\newtheorem{theorem}{Theorem}[section]
\newtheorem{proposition}[theorem]{Proposition}

\newtheorem{definition}[theorem]{Definition}

\begin{document}

%

%

\twocolumn[

\aistatstitle{The Cost of Local and Global Fairness in Federated Learning}

\aistatsauthor{ Yuying Duan \And Gelei Xu \And  Yiyu Shi \And Michael Lemmon }

\aistatsaddress{ \\ University of Notre Dame
} ]

\begin{abstract}


With the emerging application of Federated Learning (FL) in finance, hiring and healthcare, FL models are regulated to be fair, preventing disparities with respect to legally protected attributes such as race or gender.  Two concepts of fairness are important in FL: global and local fairness.  Global fairness addresses the disparity across the entire population and local fairness is concerned with the disparity within each client.  Prior fair FL frameworks have improved either global or local fairness without considering both.  Furthermore, while the majority of studies on fair FL focuses on binary settings, many real-world applications are multi-class problems. This paper proposes a framework that investigates  the minimum accuracy lost for enforcing a specified level of global and local fairness in  multi-class FL settings. Our framework leads to a simple post-processing algorithm that derives fair outcome predictors from the Bayesian optimal score functions. Experimental results show that our algorithm outperforms the current state of the art (SOTA) with regard to the accuracy-fairness tradoffs, computational and communication costs. Codes  are available at: \href{https://github.com/papersubmission678/The-cost-of-local-and-global-fairness-in-FL}{https://github.com/papersubmission678/The-cost-of-local-and-global-fairness-in-FL}.

 \end{abstract}

\section{Introduction}
\begin{table*}[t]
\caption{Comparison between our work and existing group fair FL frameworks}
\centering
\resizebox{\textwidth}{!}{%
\begin{tabular}{c|c|ccccc|c}
\toprule
 & {Frameworks} & {Binary-Class} & {Multi-Class} & {Statistical Parity} & {Equal Opportunity} & {Equalized Odds} & Approach \\
\midrule
\multirow{1}{*}{Local Fairness} 
    & \texttt{FCFL} \citep{cui2021addressing} & $\checkmark$ & $\times$  & $\checkmark$ & $\checkmark$ & $\times$ & in-processing \\
\midrule
\multirow{4}{*}{Global Fairness} 
    & \texttt{FairFed} \citep{ezzeldin2023fairfed} & $\checkmark$ & $\times$ & $\checkmark$ & $\checkmark$ & $\times$ & in-processing \\
    & \texttt{Fair-FATE} \citep{salazar2023fair} & $\checkmark$ & $\times$ & $\checkmark$ & $\checkmark$ & $\checkmark$ & in-processing \\
    & \texttt{FedFB} \citep{zeng2021improving} & $\checkmark$ & $\times$ & $\checkmark$ & $\checkmark$ & $\checkmark$ & in-processing \\
    & \texttt{Reweighing-FL} \citep{abay2020mitigating} & $\checkmark$ & $\times$ & $\checkmark$ & $\checkmark$ & $\times$ & pre-processing \\
\midrule
\multirow{2}{*}{Local and Global} 
    & \texttt{EquiFL} \citep{makhija2024achieving} & $\checkmark$ & $\times$ & $\checkmark$ & $\checkmark$ & $\times$ & in-processing \\
    & Ours & $\checkmark$ & $\checkmark$ & $\checkmark$ & $\checkmark$ & $\checkmark$ & post-processing \\
\bottomrule
\end{tabular}
}
\label{table:comparsion_with_related_work}
\end{table*}

Federated Learning (FL) \citep{mcmahan2017communication} is a distributed machine learning framework that uses data collected 
from a group of community \emph{clients} to learn a global model that can be used by all clients in the group.  The communities served by these clients are formed from sets of remote users (e.g. mobile phones) or organizations (e.g. medical clinics and hospitals) that all share some defining attribute such as a similar geographical location. FL algorithms such as \texttt{FedAvg} \citep{mcmahan2017communication} train the global model in a distributed manner by first having each client use its local data to train a local model.  This local model is then sent to the cloud server who averages these models and sends the averaged model back to the community clients who then retrain that model with their local data.  This interaction continues for  several update cycles until it converges on a global model that is agreeable to all clients.  
 FL is now the dominant framework for distributed machine learning \citep{kairouz2021advances}, particularly in smart city \citep{qolomany2020particle,pandya2023federated} and smart healthcare applications \citep{antunes2022federated,brisimi2018federated}.

Group fairness \citep{dwork2012fairness,hardt2016equality} is  a critical concern 
of FL applications such as healthcare.  
Group fairness requires that a model's decisions do not favor any particular group with 
legally protected (a.k.a sensitive) attributes such as race,  gender, or age. There are two concepts of global fairness in FL: \textit{global fairness} and \textit{local fairness}. Global fairness \citep{ezzeldin2023fairfed} addresses disparities over the entire data distribution. Local fairness \citep{cui2021addressing} addresses disparity within each local client's data distribution. Ensuring both fairness concepts is necessary in FL applications such as hospital networks. The network is legally required \citep{HealthEquity2022} to be fair to all individuals it serves,  while also ensuring fairness in each client hospital.
%


Most existing fair FL approaches focus on enforcing either global fairness \citep{ezzeldin2023fairfed, du2021fairness, abay2020mitigating} or local fairness \citep{cui2021addressing} in FL. It has been empirically observed  \citep{ezzeldin2023fairfed, du2021fairness, abay2020mitigating,cui2021addressing, makhija2024achieving}  that there is an underlying loss of accuracy when enforcing either local or global fairness. 
However, the exact minimum accuracy lost required to
enforce both fairness remains unexplored. Furthermore, most studies on fairness in FL \citep{cui2021addressing,ezzeldin2023fairfed,salazar2023fair,zeng2021improving,abay2020mitigating,makhija2024achieving,hamman2023demystifying} have focused on binary-class settings,  despite the fact that many real-world FL applications \citep{arunkumar2017multi} are multi-class problems. Most existing fair FL frameworks primarily address fairness w.r.t  Statistical Parity (SP) \citep{dwork2012fairness} and Equal Opportunity (EOp) \citep{hardt2016equality}.  Equalized Odds (EO) \citep{hardt2016equality} has been neglected. EO is important in multi-class  medical diagnosis since it ensures that individuals, regardless of their sensitive attributes, have an equal likelihood of being correctly diagnosed and receiving appropriate treatment across all diagnostic classes.


This paper provides a framework that investigates the minimum accuracy
lost in enforcing both local and global fairness (specifically Equalized Odds)  in multi-class FL.  
We formulate the problem of finding an optimal outcome predictor that satisfies fairness constraints as a convex program. The convex program's feasible set is a region under a Receiver Operating Characteristic (ROC) surface defined in \citep{landgrebe2008efficient}.
Solving the convex program is not practical due to the computational complexity of estimating the ROC surface, so we approximate that surface using a simplex, which allows us to reformulate the problem as a linear program (LP).  The training algorithm of our framework is a post-processing approach that first trains a model using $\texttt{FedAvg}$ \citep{mcmahan2017communication} and then enforces fairness through the LP. Our algorithm supports all commonly used fairness metrics, including SP, EOp and EO.
Experiments show that, compared to SOTA, our framework achieves a comparable level of fairness with a smaller accuracy cost while reducing communication costs by 20\% and computational costs by 33\%.

\section{Related Work}
\vspace{-0.2cm}
While the \textbf{cost of group fairness in centralized ML} has been well-studied  \citep{pmlr-v81-menon18a,kim2020fact,pmlr-v119-sabato20a,chzhen2019leveraging,zhao2022inherent,tang2022attainability,zhong2024intrinsic}, the notion of local fairness is  absent from the centralized settings. Existing literature thus focuses on the cost of global fairness. The accuracy lost for achieving both fairness concepts remains  unexplored.  
Our FL framework fixes this gap and studies the accuracy lost for improving both fairness concepts.

\textbf{Group fairness in FL} requires that models do not discriminate individuals based on sensitive attributes. Existing methods for global or local fairness, summarized in Table \ref{table:comparsion_with_related_work}, fall into three categories: pre-, in-, and post-processing. Pre-processing techniques \citep{abay2020mitigating} cannot address both global and local fairness. In-processing methods \citep{cui2021addressing, zeng2021improving, ezzeldin2023fairfed, salazar2023fair, makhija2024achieving} complicate the FL pipeline, lack convergence guarantees, and cannot precisely control fairness levels. Post-processing techniques \citep{duan2024postfairfederatedlearningachieving} modify pre-trained models to a fair model. Our approach is a post-processing  framework that optimally enforces both local and global fairness while allowing precise control of fairness in a multi-class setting across all common fairness metrics.

Existing \textbf{post-processing frameworks} \citep{hardt2016equality, chzhen2019leveraging, jiang2020wasserstein, denis2021fairness, gaucher2023fair, xian2023fair, xian2023efficient, zeng2024minimax, xian2024linear} primarily focuses on centralized ML and specific fairness metrics. For example, \citep{jiang2020wasserstein, gaucher2023fair, denis2021fairness, xian2023fair} only support Statistical Parity. \citep{hardt2016equality, chzhen2019leveraging} supports Equalized Odds and Equal Opportunity, but only in binary settings. Our approach differs by being a distributed framework that achieves both local and global fairness, works in binary and multi-class settings, and supports all common group fairness metrics.

\section{Preliminary Definitions}\label{sec:Preliminary}


\emph{Notational Conventions:} This paper will denote random variables using upper case letters, $X$, and lower case letters, $x$ will denote instances of those random variables.  A random variable's distribution will be denoted as $F_X$.  
Bold face lower case symbols will be reserved for vectors and
bold face upper case symbols will be reserved for matrices.

To formalize our statistical setup, we define the notion of a \textbf{client group}.

\begin{definition}
\label{def:community-group}
A \textbf{client group} consists of $K$ geographically distinct clients that we formally represent as a tuple of jointly random variables, $D = (X,A,C,Y)$ with probability distribution $F_D: \mathcal{X} \times \mathcal{A} \times
\mathcal{C} \times \mathcal{Y} \rightarrow [0,1]$.    An instance of the client group, $(x,a,c,y)$, is called an \textbf{individual} where $x \in \mathcal{X}$ is the individual's \textbf{private data vector},
$a \in \mathcal{A} = \{0,1\}$ denotes the individual's \textbf{protected sensitive attribute}, and $c \in \mathcal{C} = 
\{1,2, \ldots, K\}$ denotes which \textbf{client} the individual belongs to.   The other value, $y \in \mathcal{Y} = \{1, \cdots, N\}$ denotes the individual's \textbf{qualified outcome}.   
\end{definition}
Definition \ref{def:community-group} has a concrete interpretation in a community health application. The client group $D$ represents a city’s clinic network, where each client is a distinct neighborhood served by a single health clinic. For an individual $(x,a,c,y) \sim F_D$, the variable $x$ represents their private health data, $a$ is a protected attribute (e.g., race or gender), $c$ is their local clinic, and $y$ indicates the individual has a certain illness and is therefore 
"qualified" to access corresponding medical resources to treat that illness. 

We are interested in selecting an \textit{outcome predictor},
$\widetilde{Y}: \mathcal{X} \times \mathcal{A} \times \mathcal{C} \rightarrow \mathcal{Y}$  for client group $D$ such that
for any individual $(x,a,c,y) \sim F_D$ we have $
\widetilde{Y}(x,a,c) = y$ with a high probability.  
With these definitions and notational conventions in place, we formally define global fairness 
and local fairness. 

\begin{definition}
\label{def:multi_class_global_EO} ($\epsilon_0$-Global Fairness) Let $\epsilon_0 \in [0,1]$, the outcome predictor
$\widetilde{Y} : \mathcal{X} \times \mathcal{A} \times \mathcal{C} \rightarrow \mathcal{Y}$ for client group $D$ satisfies
\textbf{$\epsilon_0$-global equalized odds} if and only if for all $y \in \mathcal{Y}$
\begin{equation}\label{eq:global_EO}
\begin{split}
&|{\rm Pr}_D \left\{ \widetilde{Y}(X,A,C) = y\, | \, Y=y, A=1 \right\}  \\
-&{\rm Pr}_D \left\{ \widetilde{Y}(X,A,C) = y \, | \, Y=y, A=0 \right\}| \leq \epsilon_0.
\end{split}
\end{equation}
\end{definition}


\begin{definition}
\label{def:multi_class_local_EO} ($\boldsymbol{\epsilon_l}$-Local Fairness) Let $\boldsymbol{\epsilon}_l=[\epsilon_1 \cdots, \epsilon_K] \in [0,1]^K$ , the outcome predictor
$\widetilde{Y} : \mathcal{X} \times \mathcal{A} \times \mathcal{C} \rightarrow \mathcal{Y}$ satisfies
\textbf{$\boldsymbol{\epsilon}_l$-local equalized odds} if and only if for all $c \in \mathcal{C}$ and $y \in \mathcal{Y}$
\begin{equation}\label{eq:local_EO}
\begin{split}
&|{\rm Pr}_D \left\{ \widetilde{Y}(X,A,C) = y\, | \, Y=y, A=1, C=c \right\}\\ - 
&{\rm Pr}_D \left\{ \widetilde{Y}(X,A,C) = y \, | \, Y=y, A=0, C=c \right\}| \leq \epsilon_c.
\end{split}
\end{equation}
\end{definition}

Consider the \emph{true positive} of outcome predictor $\widetilde{Y}$ for class $y$, group $a$ and client $c$:
\begin{equation}
    {\rm TP}_{ac}^y(\widetilde{Y})= \mathbb{E}_{{\rm Pr}_{X|Y,A,C}}\left[\mathbf{1}(\widetilde{Y}(X,a,c)=y)\right]
\end{equation}
where $\mathbf{1}(\cdot)$ is the indicator function.

Fairness requires equal true positives between sensitive and non-sensitive groups. In the following, we use the right lower indices \( a \) and \( c \) to represent the group and community, respectively. The right upper index \( y \) denotes the class. \( (\cdot)^y_{ac} \) thus represents \( (\cdot) \) for class \( y \), group \( a \), and client \( c \).

The predictor's \emph{accuracy} is a linear combination of true positives: 
{\small
\begin{equation*} 
\begin{split} 
{\rm acc}(\widetilde{Y}) = \sum_{y,a,c \in \mathcal{Y},\mathcal{A}, \mathcal{C}}  \left[ {\rm Pr}_D(Y = y, A = a, C = c)  
\cdot {\rm TP}_{ac}^y(\widetilde{Y})\right]
\end{split} 
\end{equation*}
}
The ideal fair outcome predictor is one that achieves a true positive rate of 1 for classes, groups and clients. However, such a predictor in general is not achievable due to the noise in the data.
The following section  formally defines the feasible set of true positives for multi-class problem.

\section{Region under the ROC Surface} \label{sec: ROC_curves}

This section defines ROC surfaces used in the next section to formulate the optimal multi-class fair predictor.  To define the ROC surface, we first extend the score function in \citep{hardt2016equality} to multi-class setting. 

\begin{definition}[\small{Bayesian Optimal Score Function}] Let the function $R: \mathcal{X}\times \mathcal{A} \times \mathcal {C} \rightarrow [0,1]^N$ have components
    $R(x,a,c)=[r_1(x,a,c), r_2(x,a,c), \cdots r_N(x,a,c)]$ This function is a \textit{Bayesian Optimal Score Function} if
${\displaystyle \sum_{y=1}^N r_y(x,a,c)=1}$,
and for all $y \in \mathcal{Y}$, its components satisfy:  $r_y(x,a,c)={\rm Pr}_D(Y=y|X=x,A=a,C=c)$
\end{definition}

The Bayesian Optimal Score function maps inputs to a probability distribution over $\mathcal{Y}$, with each element representing the probability that an individual $(x,a,c)$ belongs to each class. For simplicity, we drop the input arguments and write $R(X,A,C)$ or $\widetilde{Y}(X,A,C)$ as $R$ or $\widetilde{Y}$.

\begin{definition}[Derived Outcome Predictor] \label{def:derived_outcome_preditor}
    Let $R: \mathcal{X} \times \mathcal{A} \times \mathcal{C} \rightarrow [0,1]^N$ be the Bayesian Optimal Score Function and $\boldsymbol{\theta}=[\theta_1, \theta_2, \cdots, \theta_N] \in \mathbb{R}^N_{\geq 0}$ be a given vector. The outcome predictor $\widetilde{Y}_{\boldsymbol{\theta}}: \mathcal{X}\times \mathcal{A} \times \mathcal{C} \rightarrow \mathcal{Y}$ that takes value of:
    \begin{equation} \label{eq:derived_outcome_predictor}       \widetilde{Y}_{\boldsymbol{\theta}}= y, \text{ if: } \theta_yr_y(x,a,c)= \max_{i=1}^{N} \theta_i r_i(x,a,c)
    \end{equation}
    is said to be derived from the Bayesian Optimal Score via $\boldsymbol{\theta}$. The set of all derived outcome predictors is $\{\widetilde{Y}_{\boldsymbol{\theta}}\}_{\boldsymbol{\theta} \in \mathbb{R}^N_{\geq 0}}$.
\end{definition}
A derived outcome predictor returns the highest value of the score weighted by $\boldsymbol{\theta}$. The derived outcome predictor follows the same format as in \citep{landgrebe2008efficient}. It generalizes the threshold test predictor from \citep{hardt2016equality} used in binary classification. 

The following proposition generalizes the NP Lemma \citep{neyman1933efficient} to multi-class settings.

\begin{proposition} (Appendix \ref{app:proof_NP_multi_class})\label{prop:NP_multi_class}
    Consider client $c$ and group $a$, let $\{\widetilde{Y}_{\boldsymbol{\theta}}\}_{\boldsymbol{\theta} \in \mathbb{R}^N_{\geq 0}}$ be the set of all derived outcome predictors, and let $\{\phi_g\}_{g=1}^{N-1}, \in [0,1]^N$ be a set of specified values of true positives. If $\widetilde{Y}^*: \mathcal{X}\times \mathcal{A} \times \mathcal{C} \rightarrow \mathcal{Y}$ is the solution to the following optimization problem:
    \begin{eqnarray*}
    \begin{array}{ll}
    \text{maximize} & {\rm TP}_{ac}^N (\widetilde{Y})\\
    \text{with respect to} & \widetilde{Y}: \mathcal{X} \times \mathcal{A} \times \mathcal{C} \rightarrow \mathcal{Y} \\
    \text{subject to } & \forall g \in \{1, 2, \cdots, N-1\}, \\
    &{\rm TP}_{ac}^g (\widetilde{Y}) = \phi_g
    \end{array}
    \end{eqnarray*}
    then $\widetilde{Y}^* \in \{\widetilde{Y}_{\boldsymbol{\theta}}\}_{\boldsymbol{\theta} \in \mathbb{R}^N_{\geq 0}}$.
\end{proposition}

Proposition \ref{prop:NP_multi_class} states that, given the true positive rates for the first $(N-1)$ classes, the predictor that maximizes the true positive rate for class $N$ is the \textit{derived outcome predictor}. 

Consider the vector in $[0,1]^N$ that represents the true positives for client $c$ and group $a$ of the \textit{derived outcome predictor} for a given $\boldsymbol{\theta}$. We denote the vector of true positives of the predictor $\widetilde{Y}_{\boldsymbol{\theta}}$ as $\textbf{TP}_{ac}(\widetilde{Y}_{\boldsymbol{\theta}})$:
\begin{equation*} \textbf{TP}_{ac}(\widetilde{Y}_{\boldsymbol{\theta}}) = [{\rm TP}_{ac}^1 (\widetilde{Y}_{\boldsymbol{\theta}}), \dots, {\rm TP}_{ac}^N (\widetilde{Y}_{\boldsymbol{\theta}})]^T \end{equation*} 
The ROC hypersurface for the multi-class setting is composed of the true positive vectors of all derived outcome predictors, $\{\widetilde{Y}_{\boldsymbol{\theta}}\}_{  \boldsymbol{\theta} \in \mathbb{R}^N_{\geq 0}}$.

The ROC hypersurface for client $c$ and group $a$ is thus defined as: \begin{equation} {\rm ROC}_{ac} := \{\textbf{TP}_{ac}(\widetilde{Y}_{\boldsymbol{\theta}}) : \forall \boldsymbol{\theta} \in \mathbb{R}^N_{\geq 0}\} \end{equation} 

The ROC hypersurfaces differ across clients as the data distributions vary across different clients in the FL system. We consider the region under the ROC hypersurface (RUS). The RUS is defined with respect to the separating hyperplanes that distinguishes the region under the ROC hypersurface from other regions.   
\begin{definition}
Let $\textbf{TP}_{ac}(\widetilde{Y}_{\boldsymbol{\theta}})$ be the point representing the true positive of the derived outcome predictor for a given $\boldsymbol{\theta}=[\theta_1, \cdots, \theta_N]$, the hyperplane of $\textbf{TP}_{ac}(\widetilde{Y}_{\boldsymbol{\theta}})$ is defined as:
\begin{equation*}
    \left\{\mathbf{x}\in \mathbb{R}^N|\mathbf{v}_{\boldsymbol{\theta}}^T \mathbf{x}=\mathbf{v}_{\boldsymbol{\theta}}^T \textbf{TP}_{ac}(\widetilde{Y}_{\boldsymbol{\theta}}) \right\}
\end{equation*}
where, 
\begin{equation*}
    \mathbf{v}_{\boldsymbol{\theta}}^T=[\theta_y {\rm Pr}_D(Y=y| A=a,C=c), \forall y \in \mathcal{Y} ]
\end{equation*}
\end{definition}
For any given $\boldsymbol{\theta} \in \mathbb{R}^N \geq \mathbf{0}$, it has a separating hyperplane such that the set $\{\mathbf{x}\in \mathbb{R}^N|\mathbf{v}_{\boldsymbol{\theta}}^T \mathbf{x} > \mathbf{v}_{\boldsymbol{\theta}}^T\textbf{TP}_{ac}(\widetilde{Y}_{\boldsymbol{\theta}})\}$  excludes the RUS.
\begin{definition} \label{region_under_ROC}
    The region under ${\rm ROC }_{ac}$ is:
    \begin{equation*}
        D_{ac}= \underset{\boldsymbol{\theta}\in \mathbb{R}^N _{\geq 0}}{\bigcap} 
        \left\{\mathbf{x}\in [0,1]^N|\mathbf{v}_{\boldsymbol{\theta}}^T \mathbf{x}\leq \mathbf{v}_{\boldsymbol{\theta}}^T \textbf{TP}_{ac}(\widetilde{Y}_{\boldsymbol{\theta}}) \right\}
    \end{equation*}
\end{definition}
\begin{proposition}(Appendix \ref{app:proof_convex_property})
\label{prop:convex_property}
Let $D_{ac}$ be the region defined in Def.\ref{region_under_ROC}. Then, $D_{ac}$ is a convex set.
\end{proposition}

\begin{proposition}(Appendix \ref{app: proof_true_positive_feasible})\label{prop:true_positive_feasible}
Let $D_{ac}$ be the region as defined in Def.\ref{region_under_ROC}. For any predictor $\widetilde{Y}: \mathcal{X} \times \mathcal{A} \times \mathcal{C} \rightarrow \mathcal{Y}$, let the point representing true positives of $\widetilde{Y}$ be:  
$\textbf{TP}_{ac}(\widetilde{Y}) = [{\rm TP}^{1}_{ac}(\widetilde{Y}), \cdots, {\rm TP}^{N}_{ac}(\widetilde{Y})]$.
 Then, $ \textbf{TP}_{ac}(\widetilde{Y})$ lies in $D_{ac}$.
\end{proposition}

Proposition \ref{prop:true_positive_feasible}
 asserts that for group $a$ in client $c$, any outcome predictor that is a map $\widetilde{Y}: \mathcal{X} \times \mathcal{A} \times \mathcal{C} \rightarrow \mathcal{Y}$, has its true positives lying in $D_{ac}$. Thus, $D_{ac}$ is the set representing the feasible region of true positives.

\section{The Minimum Cost of Fairness}
This section presents a framework that returns an optimal fair outcome predictor. The framework is implemented by solving an LP with polynomial computational complexity.  It is applicable to large-scale distributed learning systems.

 Consider the loss function $\ell : \mathcal{Y} \times \mathcal{Y} \rightarrow \{0,1\}$ that takes values 
\begin{eqnarray}
\ell(\widetilde{y},y) =
\mathbf{1}(\widetilde{y} \neq y)
\label{eq:loss}
\end{eqnarray}
 for any $\widetilde{y}, y \in \mathcal{Y}$, where $\mathbf{1}(\cdot)$ is the indicator function. Let $\boldsymbol{\epsilon}=[\epsilon_0, \epsilon_1,\epsilon_2,\cdots, \epsilon_{K}]\in [0,1]^{K+1}$ be a pre-specified fairness level.
Any predictor
$\widetilde{Y} : \mathcal{X} \times \mathcal{A} \times \mathcal{C} \rightarrow \mathcal{Y}$ that satisfies the following optimization problem is an $\textbf{$\boldsymbol{\epsilon}$-fair optimal outcome predictor}$ 
\begin{eqnarray}
\begin{array}{ll}
\text{minimize} & \mathbb{E}_{D} \left[ \ell(\widetilde{Y}(X,A,C),Y) \right] \\
\text{with respect to} & \widetilde{Y}: \mathcal{X} \times \mathcal{A} \times \mathcal{C} \rightarrow \mathcal{Y} \\
\text{subject to }
& \text{eq.} (\ref{eq:global_EO}) \text{ and } \text{eq.} (\ref{eq:local_EO})
\end{array}
\label{eq:optimization-problem}
\end{eqnarray}

\label{sec:lp}
Optimization \eqref{eq:optimization-problem} selects an outcome predictor that satisfies the specified local and global fairness constraints while minimizing the classifier's error rate.  Optimization \eqref{eq:optimization-problem} can be recast as the following convex program.

For notational convenience, let $z_{ac}^y$ be the variables for the outcome predictor $\widetilde{Y}$:
\begin{eqnarray*}
    z_{ac}^y= TP_{ac}^y(\widetilde{Y})= {\rm Pr}_D (\widetilde{Y}=y|Y=y, A=a, C=c)
\end{eqnarray*}
that represents the true positives of $\widetilde{Y}$ for class $y$, group $a$ and client $c$. The following proposition asserts that if $\{z_{ac}^y \}_{\mathcal{Y}, \mathcal{A}, \mathcal{C}}$ satisfies the following convex program, then $\widetilde{Y}$ is a $\epsilon-$fair optimal outcome predictor.

\begin{proposition} (Appendix \ref{app:proof_convex_program})
\label{prop:convex_program}
Let the vector $\mathbf{z} \in \mathbb{R}^{2 N  K}$:
\begin{eqnarray*}
\mathbf{z}^T &=& \left[ \begin{array}{ccccccc}
\mathbf{z}_{01}^T & \mathbf{z}_{11}^T & \mathbf{z}_{02}^T& \mathbf{z}_{12}^T \cdots & \mathbf{z}^T_{0K}& \mathbf{z}^T_{1K} \end{array} \right]
\end{eqnarray*}
where $\mathbf{z}_{ac}$ are true positives of all classes: $y \in \mathcal{Y}$ for group $a$ in client $c$:
\begin{eqnarray*}
    \mathbf{z}_{ac}^T &=& \left[ \begin{array}{cccccc}
  z_{ac}^{1}& z_{ac}^{2}& z_{ac}^{3}& \cdots &z_{ac}^{N} \end{array}\right] \in \mathbb{R}^N
\end{eqnarray*}
satisfy the following convex program
\begin{eqnarray}\label{eq:convex_program}
\begin{array}{ll}
\mbox{minimize:} &\mathbf{c}^T \mathbf{z} \\
\mbox{with respect to:}& \mathbf{z}  \in \mathbb{R}^{2NK} \\
\mbox{subject to:} & -\mathbf{b}\leq \mathbf{Az} \leq \mathbf{b} \\
& \mathbf{z}_{ac} \in D_{ac}, \forall a \in \mathcal{A}, c \in \mathcal{C}
\end{array}
\end{eqnarray}
then, the outcome predictor $\widetilde{Y}:\mathcal{X} \times \mathcal{A} \times \mathcal{C} \rightarrow \mathcal{Y}$ that satisfies eq. \eqref{eq:z_ac_y_value} for all $ y \in \mathcal{Y}, a\in \mathcal{A}, c\in \mathcal{C}$
\begin{equation} \label{eq:z_ac_y_value}
     {\rm Pr}(\widetilde{Y}=y|Y=y, A=a,C=c)=z_{ac}^{y}
\end{equation}
is an $\boldsymbol{\epsilon}$-fair optimal outcome predictor.The optimal accuracy for a $\epsilon$-fair outcome predictor is $-\mathbf{c}^T\mathbf{z}.$

\end{proposition}

The parameters of \eqref{eq:convex_program}, $\mathbf{A} \in \mathbb{R}^{N(K+1) \times 2NK}$, $\mathbf{c} \in \mathbb{R}^{2NK}$ and $\mathbf{b}\in \mathbb{R}^{N(K+1)}$ are detailed in Appendix \ref{app: parameters of LP}.  $D_{ac}$ is the region under the ROC hypersurface as defined in  Def. \ref{region_under_ROC}.
Optimization \eqref{eq:convex_program} is a convex program since $D_{ac}$ is a convex set (Proposition \ref{prop:convex_property}). 
The first $N$ inequalities of $ \mathbf{-b}\leq\mathbf{A}\mathbf{z}\leq \mathbf{b}$ represent global fairness constraints and the last $(NK)$ inequalities represent local fairness constraints.

Since $D_{ac}$ is the feasible region for the true positives of all outcome predictors (Proposition \ref{prop:true_positive_feasible}), the  predictor $\widetilde{Y}$ is fair and has optimal accuracy. The convex program \eqref{eq:convex_program} thus presents the \textit{inherent trade-off} between global fairness, local fairness and accuracy and characterizes the minimum accuracy lost in enforcing specified level of global and local fairness.

The computational complexity of generating the convex set $D_{ac}$  increases exponentially w.r.t the number of classes \citep{landgrebe2008efficient}. Instead of solving the convex program \eqref{eq:convex_program} directly, our implementation approximates the $D_{ac}$ using a simplex $\widehat{D_{ac}}$, reformulating the problem as a linear program.

Let us consider the \textit{derived outcome predictor} $ \widetilde{Y}_{\boldsymbol{\theta}}$ in Def \ref{def:derived_outcome_preditor} with $\boldsymbol{\theta}=\mathbf{1}$.
$\widetilde{Y}_{\mathbf{1}}: \mathcal{X} \times \mathcal{A} \times \mathcal{C} \rightarrow \mathcal{Y}$ takes values of:
\begin{equation}\label{eq:optimal_predictor}
    \widetilde{Y}_\mathbf{1}(x,a,c) = \arg\max_{y \in \mathcal{Y}} r_y(x,a,c)
\end{equation}
The point representing true positives for $\widetilde{Y}_{\mathbf{1}}$ is: $    \textbf{TP}_{ac}(\widetilde{Y}_{\mathbf{1}}) = [{\rm TP}_{ac}^1 (\widetilde{Y}_{\boldsymbol{1}}),\cdots, {\rm TP}_{ac}^N (\widetilde{Y}_{\boldsymbol{1}})]$. The simplex we use to approximate the convex set $D_{ac}$ is the one whose vertices are  $\{\mathbf{e}_y\}_{ \forall y \in \mathcal{Y}}$ and $\textbf{TP}_{ac}(\widetilde{Y}_{\mathbf{1}})$:
\small{\begin{equation*}
    \widehat{D_{ac}}= \{f_0\textbf{TP}_{ac}(\widetilde{Y}_{\mathbf{1}})+ \sum_{y=1}^{N} f_y\mathbf{e}_y| \sum_{i=0}^{N} f_i=1\text{ and }\forall i, f_i\geq 0 \}
\end{equation*}}
where, $ \mathbf{e}_y$ are elementary basis vectors. We choose \( \widehat{D_{ac}} \) to be an inner approximation to \( D_{ac} \) $(\widehat{D_{ac}} \subset D_{ac})$. \( \widetilde{Y}_{\mathbf{1}} \) is the predictor that maximizes the accuracy for the distribution \( D \) (Appendix \ref{app:optimal_of_y_1}).


The polytope $\widehat{D_{ac}}$ can be represented using $(N+1)$ inequalities. Any point $\mathbf{u} \in \mathbb{R}^N$ that lies in $\widehat{D_{ac}}$ must satisfy:
\begin{equation}
    \mathbf{K}_{ac}\mathbf{u}\leq \mathbf{l}_{ac}
\end{equation}
where $\mathbf{K}_{ac} \in \mathbb{R}^{(N+1) \times N}$ and $\mathbf{l}_{ac} \in \mathbb{R}^{(N+1)}$ are detailed in Appendix \ref{app:convex_hull}.

We now reformulate the problem \eqref{eq:convex_program} as LP \eqref{eq: lp}:
\begin{eqnarray}\label{eq: lp}
\begin{array}{ll}
\mbox{minimize:} &\mathbf{c}^T \mathbf{z} \\
\mbox{with respect to:}& \mathbf{z}  \in \mathbb{R}^{2NK} \\
\mbox{subject to:} & -\mathbf{b} \leq \mathbf{Az} \leq \mathbf{b} \\
&\mathbf{K}_{ac}\mathbf{z}_{ac} \leq \mathbf{l}_{ac}, \forall a \in \mathcal{A}, c \in \mathcal{C}
\end{array}
\end{eqnarray}
The LP has the same parameters as the convex program, except for the final linear constraint: $\mathbf{K_{ac}}\mathbf{z}_{ac} \leq \mathbf{l}_{ac}$. LP \eqref{eq: lp} can be extended to other fairness metrics (see Appendix \ref{app:extension_to_other_fairness}
).


{\small
\begin{algorithm*}[t]
    \caption{Fair Outcome Predictor}
    \label{alg:1}
        \textbf{Input:} The  outcome predictor: $\widetilde{Y}_{\mathbf{1}}: \mathcal{X}\times \mathcal{A} \times \mathcal{C}\rightarrow \mathcal{Y}$, the client $c$'s  $\boldsymbol{\beta}_{ac}, \forall a \in \mathcal{A}$
        
        \quad \quad $\boldsymbol{\beta}_{ac}^T=[\beta_{ac}^0, \beta_{ac}^{1}, \beta_{ac}^{2}, \cdots \beta_{ac}^{N}]\in [0, 1]^{N+1}$
         
    \textbf{Output:}  Fair outcome predictor $\widetilde{Y}_{\boldsymbol{\beta}_{ac}}: \mathcal{X}\times \mathcal{A} \times \mathcal{C}\rightarrow \mathcal{Y}$     


        \quad 1. randomly sample $s \sim U(0,1)$, the uniform distribution between (0,1)

        \quad 2. Construct $\widetilde{Y}_{\boldsymbol{\beta}_{ac}}(x,a,c)$ as

        \quad \quad $\widetilde{Y}_{\boldsymbol{\beta}_{ac}}(x,a,c)=\left\{\begin{array}{l}
        \widetilde{Y}_{\mathbf{1}}(x,a,c),  \text { if } s \leq \beta_{ac}^0\\ [0.12cm]
        1,  \text { if } \beta_{ac}^0 < s\leq \sum_{i=0}^{i=1}\beta_{ac}^{i}  \\[0.12cm]
        2,  \text { if } \sum_{i=0}^{i=1}\beta_{ac}^{i} < s\leq \sum_{i=0}^{i=2}\beta_{ac}^{i}  \\[0.12cm]
        3,  \text { if } \sum_{i=0}^{i=2}\beta_{ac}^{i} < s\leq \sum_{i=0}^{i=3}\beta_{ac}^{i}  \\
        \cdots\\
        N,  \text { if } \beta_{ac}+\sum_{i=0}^{i=N-1}\beta_{ac}^{i}< s\leq \beta_{ac}+\sum_{i=0}^{i=N}\beta_{ac}^{i}
        \end{array}\right.$

        \textbf{return} $\widetilde{Y}_{\boldsymbol{\beta}_{ac}}$

\end{algorithm*}
}
LP \eqref{eq: lp} identifies the true positives for the fair outcome predictor. The next step is to find a classifier that achieves these true positives. The following proposition demonstrates the existence and uniqueness of the fair outcome predictor given the LP solution.

\begin{proposition} (Appendix \ref{app:proof_uniqueness})\label{prop:uniqueness}
   Let $\mathbf{z} \in \mathbb{R}^{2NK}$ be the solution of the LP (\ref{eq: lp}) 
   \begin{eqnarray*}
    \mathbf{z}^T &=& \left[ \begin{array}{ccccccc}
    \mathbf{z}_{01}^T & \mathbf{z}_{11}^T & \mathbf{z}_{02}^T& \mathbf{z}_{12}^T \cdots & \mathbf{z}^T_{0K}& \mathbf{z}^T_{1K} \end{array} \right]
    \end{eqnarray*} and ${\rm TP}_{ac}^y (\widetilde{Y}_{\boldsymbol{1}})$ be the true positive of the predictor  defined in Eq. \eqref{eq:optimal_predictor}.
   For all $a \in \mathcal{A}, c\in \mathcal{C}$, let ${\boldsymbol{\beta}_{ac}}=[\beta^0_{ac}, \beta^{1}_{ac}, \cdots, \beta^{N}_{ac}]$ be the solution of the following linear algebraic equation (LAE),
   \begin{equation}
       \label{eq:ugh-ugh}
       \mathbf{G}_{ac} \boldsymbol{\beta}_{ac}= \boldsymbol{\gamma}_{ac}
   \end{equation}
   where, the parameter $\mathbf{G}_{ac} \in \mathbb{R}^{(N+1)\times (N+1)}, \boldsymbol{\gamma}_{ac} \in \mathbb{R}^{N+1}$, are detailed in Appendix \ref{app:para_LAE}. Then. the predictor $\widetilde{Y}_{\boldsymbol{\beta}_{ac}}$ that takes value,
         \small
        \begin{equation} \label{eq:fair_outcome_predictor}
    \widetilde{Y}_{\boldsymbol{\beta}_{ac}}(x,a,c)=
        \left\{
      \begin{array}{ll}
      \widetilde{Y}_{\mathbf{1}}(x,a,c),  \quad \text{ with the probability }  \beta_{ac}^0 \\[0.2cm]
        y , \quad \text{with the probability } \beta^y_{ac},  \forall y \in \mathcal{Y}\\[0.2cm]  
      \end{array}
      \right.
    \end{equation}
    \normalsize
    is a fair outcome predictor. There always exists a unique set of parameters $\{\boldsymbol{\beta}_{ac}\}_{\mathcal{A},\mathcal{C}}$, where $\boldsymbol{\beta}_{ac} \in [0,1]^{N+1}$ and $|\boldsymbol{\beta}_{ac}|_{\ell_1}=1$ that satisfies the LAE.
\end{proposition}

Proposition \ref{prop:uniqueness}
 gives the fair outcome predictor from the LP solution.  Combining the LP \eqref{eq: lp} and LAE \eqref{eq:ugh-ugh},
our framework first solves the LP (\ref{eq: lp}) that identifies the true positives for the fair outcome predictor. Then, the fair outcome predictor \eqref{eq:fair_outcome_predictor} can be uniquely determined by solving the LAE \eqref{eq:ugh-ugh}.



\section{Training Fair Predictor in FL} \label{sec: training_steps}
{\vspace{-0.15cm}}
This section shows how LP (\ref{eq: lp}) and LAE \eqref{eq:ugh-ugh} can be used within a FL framework to construct a fair outcome predictor. 
The training steps of our framework are enumerated below: 

\underline {Step 1:} \textbf{Training an Optimal Score Function Using \texttt{FedAvg}}: The server and clients collaboratively train a Bayesian optimal score function  $R: \mathcal{X}\times \mathcal{A}\times \mathcal{C} \rightarrow [0,1]^N$ that minimizes the categorical cross-entropy loss using the \texttt{FedAvg} algorithm \citep{mcmahan2017communication}. 

 The \texttt{FedAvg} algorithm converges to the Bayesian Optimal Score at a rate of $\mathcal{O}(1/T)$, where $T$ is the number of communication round \citep{li2019convergence}. 

\underline {Step 2:} \textbf{Computing Client Statistics}: Each local client generates the output $\widetilde{Y}_{\mathbf{1}}(X,A,C)$ using Eq. \ref{eq:optimal_predictor} , computes the following empirical statistics (\ref{eq: est_statistics}) and transmits these probabilities to the global server.
{\small
\begin{equation} \label{eq: est_statistics}
    \begin{split}
        &{\rm Pr}_D (\widetilde{Y}_{\mathbf{1}}=y, Y=y, A=a \mid C=c)\\= &\frac{\text{ $\#$ of samples with ($\widetilde{Y}_{\mathbf{1}}=y, Y=y, A=a$) in client $c$}}{\text{$\#$ of samples in client c}}\\
        &{\rm Pr}_D (Y=y, A=a \mid C=c)\\= &\frac{\text{ $\#$ of samples with ($Y=y, A=a$) in client $c$}}{\text{$\#$ of samples in client c}}   \end{split}
\end{equation}}
In this step, the client transmits statistics to the serve  the individual's private data  $x \in \mathcal{X}$, such as health profiles remains stored locally.

\underline {Step 3:} \textbf{Constructing and Solving the Linear Program}: The global server computes the parameters of the LP \eqref{eq: lp} using the statistics sent by the clients.
The parameters of the LP, as detailed in Appendix \ref{app: parameters of LP}, $p_{ac}^{y}, \alpha_{a}^{y}, {{\rm TP}_{ac}^{y}}(\widetilde{Y}_{\mathbf{1}})$  are computed as:
{ \small \begin{equation}
\label{eq:parameters}
    \begin{split}
        p_{ac}^{y}&= {\rm Pr}_D(Y=y,A=a|C=c)\cdot p_c, 
        \quad \alpha_{a}^{y} = \sum_{c \in \mathcal{C}}p_{ac}^{y}\\
        {{\rm TP}_{ac}^{y}}(\widetilde{Y}_{\mathbf{1}})&= \frac{{\rm Pr}_D(\widetilde{Y}_{\mathbf{1}}=y, Y=y, A=a|C=c)}{{\rm Pr}_D( Y=y, A=a|C=c)}\\  
        \text{with, }p_c &= \frac{\text{$\#$ of samples in client $c$}}{\text{$\#$ of total samples}}
    \end{split}
\end{equation}
\small
 ${\small{\rm Pr}_D({\widetilde{Y}_{\mathbf{1}}=y, Y=y, A=a \mid C=c}), {\rm Pr}_D({Y=y, A=a \mid C=c}) }$ 
 \normalsize
 are the statistics transmitted from local clients.
 
Using the above parameters, the global server constructs the linear program (\ref{eq: lp}), finds the minimizer $
\mathbf{z}^T = \left[ \begin{array}{ccccccc}
\mathbf{z}_{01}^T & \mathbf{z}_{11}^T & \mathbf{z}_{02}^T& \mathbf{z}_{12}^T \cdots & \mathbf{z}^T_{0K}& \mathbf{z}^T_{1K} \end{array} \right]$

and then sends the corresponding minimizer $\mathbf{z}_{0c}^T, \mathbf{z}_{1c}^{T}$ to client $c$, where $c = 1, 2, \cdots, K$.

\underline {Step 4:} \textbf{Solving the LAE:} Local clients compute the parameters of the LAE \eqref{eq:ugh-ugh}. The parameter ${{\rm TP}_{ac}^{y}}(\widetilde{Y}_{\mathbf{1}})$ is computed in the same way as in equation (\ref{eq:parameters}), and $\mathbf{z}_{ac}$ is sent by the server to the clients. The local clients then construct and solve the LAE \eqref{eq:ugh-ugh}. The solution, $\boldsymbol{\beta}_{ac}$, is used in Algorithm \ref{alg:1} to make fair predictions.

Algorithm \ref{alg:1} provides a client-dependent randomized function that returns a model, which takes the value $\widetilde{Y}_{\mathbf{1}}$ with probability $\beta_{ac}^{0}$ and $y$ with probability $\beta_{ac}^{y}$.
\section{Experiments}
{\begin{table*}[h]
\caption{Accuracy, local and global disparity of all algorithms. All measurement is computed over 5 runs. }
\centering
\fontsize{9pt}{9pt}\selectfont
{%
\begin{tabular}{c|c c c  c }
\toprule
\textbf{Dataset} & \textbf{Frameworks} & \textbf{Local Disparity $(\downarrow)$} & \textbf{Global Disparity $(\downarrow)$} & \textbf{Avg-Acc $(\%) (\uparrow)$} \\
\midrule
\multirow{8}{*}{Adult (SP)} & \textit{FedAvg} & $0.287 \pm 0.037$ & $0.226 \pm 0.033$ & $\mathbf{84.5 \pm 0.6}$ \\ 
                       & \textit{FCFL} & $0.052 \pm 0.011$  & $0.055 \pm 0.020$ & $79.5 \pm 0.7$  \\ 
                       & \textit{FairFed} & $\mathbf{0.031 \pm 0.022}$ & $0.030 \pm 0.009$  & $78.4 \pm 2.1$ \\ 
                       & \textit{Fair-FATE} &$0.041 \pm 0.039 $ & $0.037 \pm 0.098$  &  $79.7 \pm 3.4$\\ 
                       & \textit{EquiFL} &$0.053 \pm 0.064$  & $0.039 \pm 0.062$  & $77.0 \pm 2.7$  \\ 
                       & \textit{Ours} & $0.039 \pm 0.012$  & $\mathbf{ 0.003 \pm 0.001}$  & $81.0 \pm 0.3$\\ 
                       & \textit{Ours (only local)} & $0.044 \pm 0.015$  & $ 0.023 \pm 0.013$  & $81.0 \pm 0.2$ \\ 
                       & \textit{Ours (only global)} & $0.155 \pm 0.001$ & $\mathbf{0.003 \pm 0.002}$ &$81.0 \pm 0.3$  \\ 
\midrule
\multirow{8}{*}{PublicCoverage (EOp)} & \textit{FedAvg} & $0.103 \pm 0.003 $ & $0.062 \pm 0.004$ &$\mathbf{77.8  \pm 0.4}$ \\
                       & \textit{FCFL} &$0.051 \pm 0.020$  & $0.032 \pm 0.02$ & $76.5 \pm 0.4$  \\ 
                       & \textit{FairFed} &  $ 0.091\pm 0.031 $& $ 0.003\pm 0.001$  & $ {76.4\pm 0.4} $ \\ 
                       & \textit{Fair-FATE} &$ 0.077 \pm 0.050$ &$0.049 \pm 0.066$  &$ 69.9 \pm 6.1$ \\ 
                       & \textit{EquiFL} &$0.051 \pm0.001 $  & $ 0.042\pm 0.002 $  & $ 76.5\pm 0.1 $  \\ 
                       & \textit{Ours} & $ \mathbf{0.041 \pm 0.001 }$  & $ \mathbf{0.001 \pm0.002} $   & $ 76.4\pm 0.2$\\ 
                       & \textit{Ours (only local)} & $0.048 \pm 0.002$  & $ 0.020 \pm 0.001$  &$ 76.5 \pm 0.6$ \\ 
                       & \textit{Ours (only global)} & $0.163 \pm 0.001$ & $\mathbf{0.001 \pm 0.000}$ &$76.6 \pm 0.0$  \\ 
\midrule
\multirow{4}{*}{HM10000 (EO)} & \textit{FedAvg} & $ 0.421\pm 0.040 $ &  $ 0.198\pm 0.024 $ & $\mathbf{ 75.4\pm 0.6} $ \\ 
                       & \textit{Ours} & $\mathbf{ 0.113\pm 0.021}$  & $0.056  \pm 0.025$  & $ 69.4 \pm 0.2$\\ 
                       & \textit{Ours (only local)} & $0.116\pm 0.050$  &  $ 0.161 \pm 0.031$  & $ 69.5\pm 0.2$ \\
                       & \textit{Ours (only global)} & $ 0.241\pm 0.020$ & $\mathbf{0.048 \pm 0.017}$ &$ 74.9 \pm 0.1$  \\ 
\bottomrule
\end{tabular}%
}
\label{table:acc_with_baselines}
\end{table*}}

\begin{table*}[ht]
\caption{The number of communication round and total time for convergence for all algorithms }
\centering
\resizebox{\textwidth}{!}
{%
\begin{tabular}{c|  c c c c c|c c c c c}
\toprule
& \multicolumn{5}{c|}{Adult}& \multicolumn{5}{c}{PublicCoverage}\\
\midrule
Frameworks & \textit{FCFL} & \textit{FairFed} & \textit{Fair-FATE} & \textit{EquiFL} & \textit{Ours} & \textit{FCFL} & \textit{FairFed} & \textit{Fair-FATE} & \textit{EquiFL} & \textit{Ours} \\ \midrule
\# of convergence rounds & $>1000$& $\approx 15$ &
$\approx 12$& $\approx 30$ & $\mathbf{\approx 10}$ & $ >1000$&$\approx 15$&$\approx 15$&$\approx 30$& $\mathbf{\approx 10}$\\ 
Total time for convergence (s) $\approx$ &220.1&27.9 &40.6&54.3 & $\mathbf{18.8}$ & 1080.0 & 484.8& 563.8 & 852.2 & $\mathbf{227.8}$\\
\bottomrule
\end{tabular}%
}
\label{table:one_round_time}
\end{table*}

\begin{figure*}[h]
    \centering
    \begin{subfigure}[t]{0.33\textwidth}
        \centering
        \includegraphics[width=\textwidth]{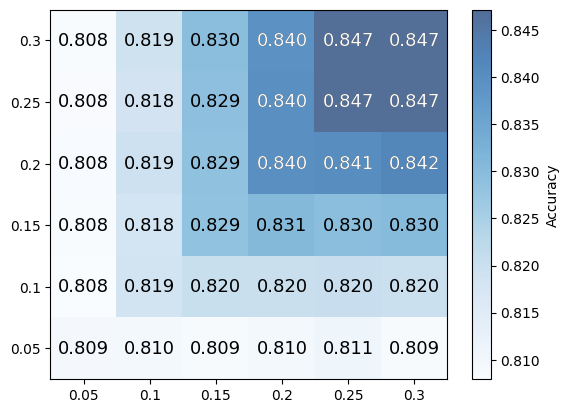}
        \caption{Adult}
        \label{fig:sub1}
    \end{subfigure}%
    \begin{subfigure}[t]{0.33\textwidth}
        \centering
        \includegraphics[width=\textwidth]{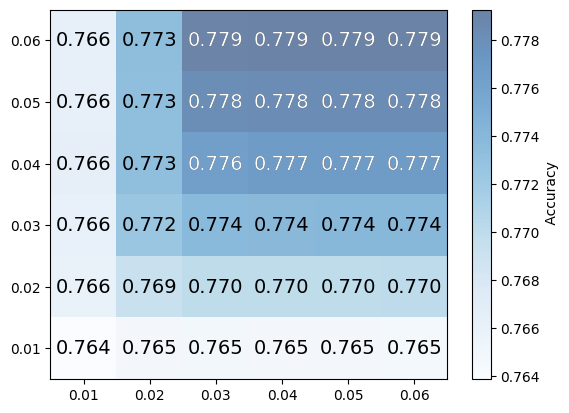}
        \caption{ACSPublicCoverage}
        \label{fig:sub2}
    \end{subfigure}%
    \begin{subfigure}[t]{0.33\textwidth}
        \centering
        \includegraphics[width=\textwidth]{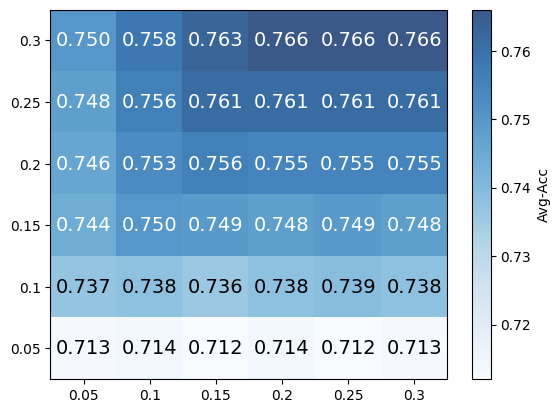}
        \caption{HM10000}
        \label{fig:sub3}
    \end{subfigure}
\caption{The accuracy under different levels of local fairness (y-axis) and global fairness (x-axis). The numbers inside the blocks are the accuracy, with each value being the average of 10 runs.}
    \label{fig:three_images}
\end{figure*}


\normalsize
We conduct experiments on real-world datasets.  Results  shows that our framework significantly reduces both local and global disparity, achieving competitive accuracy while ensuring fairness compared to SOTA.  Furthermore, our framework outperforms SOTA in communication efficiency and computational cost.
\textbf{Dataset.} We evaluate classification tasks on three datasets: Adult Dataset, ACSPublicCoverage Dataset and HM10000 Dataset.

(1) \textbf{Adult} \citep{asuncion2007uci} predicts whether an individual earns over 50K/year. Following \citep{cui2021addressing}, the dataset is split into two clients based on PhD status, with gender as the sensitive attribute. The fairness metric is Statistical Parity (SP), ensuring equal likelihood of earning over 50K/year for males and females.

(2) \textbf{ACSPublicCoverage} \citep{ding2021retiring} predicts whether an individual has public health insurance. It is divided by 50 U.S. states, with race (white/non-white) as the sensitive attribute. The fairness metric is Equal Opportunity (EOp), ensuring equal probability of coverage for eligible individuals, regardless of race.

(3) \textbf{HM10000} \citep{tschandl2018ham10000} includes 10,000 dermatoscopic images and we use it to predict 4 diagnostic class. The dataset is split into five age-based communities, with gender as the sensitive attribute. The fairness metric is Equalized Odds (EO), ensuring equal likelihood of correct diagnosis and treatment across all classes, regardless of gender.

\textbf{Evaluation.} 
We assess the model from three perspectives:  model performance, local fairness and global fairness. (1) Model performance is measured by  accuracy over all samples. For different datasets, we enforce fairness w.r.t different metrics and measure it accordingly. We use SP Difference (difference in positive rate between the sensitive and the non-sensitive) for Adult, EOp Difference (difference in true positives for class 1) for ACSPublicCoverage and EO Difference (maximum difference in true positives across all classes) for HM10000. (2) Local fairness is  measured by average disparity within each client. (3) Global fairness, measured by disparity across the entire dataset.

\textbf{Our Implementation and Baselines.}
We configure our implementation to enforce only local fairness ($\epsilon_0=1, \epsilon_c=0.01$), only global fairness ($\epsilon_0=0.01, \epsilon_c=1$)  and both ($\epsilon_0=0.01, \epsilon_c=0.01$) by adjusting $\epsilon_0$ and $\epsilon_c$ in the LP \eqref{eq: lp}. We compare ours to existing baelines that are designed to enforce local fairness: $\texttt{FCFL}$ \citep{cui2021addressing}, global fairness: $\texttt{Fair-Fate}$ \citep{salazar2023fair} $\texttt{FairFed}$ \citep{ezzeldin2023fairfed} and both $\texttt{EquiFL}$ \citep{makhija2024achieving}. These baselines focus on  a binary setting and most of them cannot be extended to enforce EO in multi-class. We compare ours with them in two binary classification tasks: Adult and ACSPublicCoverage.
Full experimental details, including models, baselines and hyperparameters are detailed in Appendix \ref{app:exp_details}. 
We summarize the main experimental results in the following:

\textbf{Fairness of Our Implementation.} 
Table \ref{table:acc_with_baselines} shows accuracy, local disparity, and global disparity for three datasets. Compared to $\texttt{Fedavg}$, our method reduces local disparity by 86.4\% (0.287 $\rightarrow$ 0.044) for Adult, 60.2\% (0.103 $\rightarrow$ 0.041) for ACSPublicCoverage, and 73.2\% (0.421 $\rightarrow$ 0.113) for HM10000. Global disparity decreases by 89.8\% (0.226 $\rightarrow$ 0.023) for Adult, 98.4\% (0.062 $\rightarrow$ 0.001) for ACSPublicCoverage, and 71.7\% (0.198 $\rightarrow$ 0.056) for HM10000. The statistical  fluctuation of our implementation is less than 1\% relative to the reduction. These results demonstrate that our implementation significantly improves both fairness concepts in FL.

Comparing ours with local constraints and Ours with global constraints in Table \ref{table:acc_with_baselines}, enforcing local fairness also reduces global disparity by 89.8\% for Adult, 67.7\% for ACSPublicCoverage, and 20\% for HM10000. Conversely, enforcing global fairness reduces local disparity by 45\% for Adult and 42\% for HM10000 but increases it by 58\% for ACSPublicCoverage. It shows that enforcing one type of fairness impacts the other.

\textbf{Flexibility of Adjusting the Level of Fairness.}
Fig. \ref{fig:three_images} shows the accuracy under varying levels of local fairness (y-axis) and global fairness (x-axis). It demonstrates that  by changing the parameters in LP \eqref{eq: lp}, the trade-off between accuracy and fairness can be flexibly adjusted. The ratio of accuracy lost for local disparity reduction to global disparity reduction is 0.98, indicating that two reductions have a close impact on accuracy. For the ACSPublicCoverage, the ratio is 1.12, indicating that reducing local disparity has a slightly higher impact on accuracy. In the HM10000 dataset, the accuracy lost of reducing local disparity is significantly higher than reducing global disparity. These results shows that the impact of local and global fairness on accuracy varies across different datasets.

\textbf{Accuracy Cost of Fairness.} Table \ref{table:acc_with_baselines} shows the accuracy lost for our implementation of  achieving both fairness objectives is 3.5\% (84.5 $\rightarrow$ 81) for Adult, 1.4\% (77.8 $\rightarrow$ 76.5) for PublicCoverage and 6\% (75.4 $\rightarrow$ 69.4) for HM10000.  Compared to other baselines,  our framework achieves comparable or better fairness levels, the accuracy cost of ours is $0.5\%- 4\%$ less for the Adult dataset and $0\%- 6\%$ less for PublicCoverage and HM10000. This supports the assertion that our approach has the smaller or comparable accuracy lost for improving fairness.

\textbf{Communication and Computational Cost.}
We use the number of communication rounds required as a measure of communication efficiency and the time to convergence as a measure of computational cost. The computation time is obtained on a 16-Core 4.00 GHz AMD Pro 5955WX Processor.
The results in Table \ref{table:one_round_time} show that the number of communication rounds for our method is 20\% smaller  for the Adult and 33\% smaller for PublicCoverage compared with baselines that has smallest communication round ($\texttt{Fair-Fate}$) . The computational time is 32.6\% less than for Adult and 53\% less for PubCoverage than the fastest baseline ($\texttt{FairFed}$). These results show that our implementation improves fairness in FL with significantly lower communication and computational cost.

\begin{figure*}[t]
    \centering
    \begin{subfigure}[t]{0.33\textwidth}
        \centering
        \includegraphics[width=\textwidth]{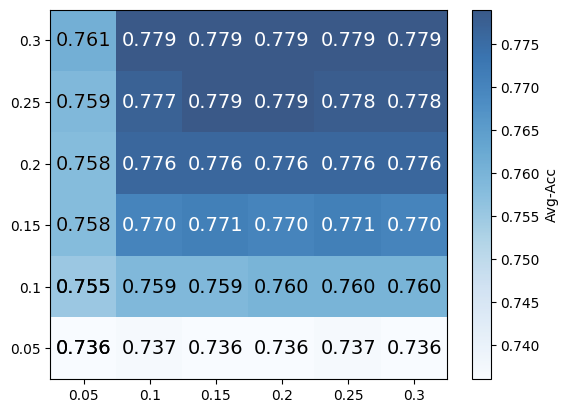}
        \caption{Scenario 1}
        \label{fig:sub1}
    \end{subfigure}%
    \begin{subfigure}[t]{0.33\textwidth}
        \centering
        \includegraphics[width=\textwidth]{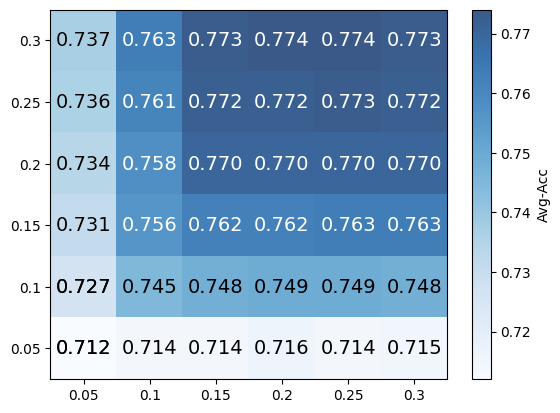}
        \caption{Scenario 2}
        \label{fig:sub2}
    \end{subfigure}%
    \begin{subfigure}[t]{0.33\textwidth}
        \centering
        \includegraphics[width=\textwidth]{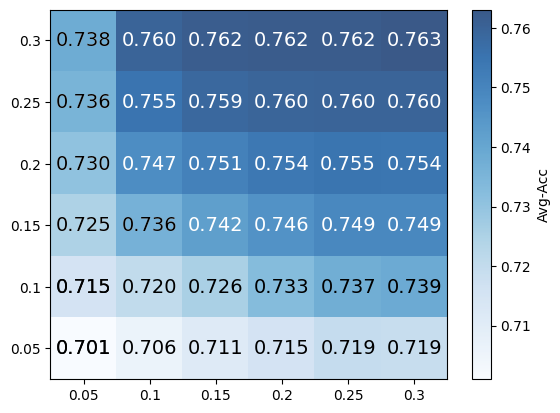}
        \caption{Scenario 3}
        \label{fig:sub3}
    \end{subfigure}
\caption{The accuracy under different levels of local (y-axis) and global fairness (x-axis) for different data heterogeneity setup of HM10000. (each value is the average of 10 runs.)}
    \label{fig:impact_data_noniid}
\end{figure*}

\textbf{Impact of Data Heterogeneity}
We investigate how data heterogeneity impacts the trade-off between global fairness, local fairness and accuracy. Data heterogeneity in FL refers to the scenarios that the data across clients in FL are non-i.i.d. Experiments were conducted on the HM10000 dataset, where the client makeup with respect to the sensitive attribute was varied. The results for three different data heterogeneity scenarios are shown in Fig. \ref{fig:impact_data_noniid}.
In Scenario 1, the proportion of sensitive (female) samples across the 5 clients is $[0.4, 0.4, 0.4, 0.4, 0.4]$. The the sensitive attribute is i.i.d. across clients. In Scenario 2, the proportion of sensitive samples of each client is $[0.2, 0.3, 0.4, 0.5, 0.6]$. In Scenario 3, it is $[0.1, 0.3, 0.5, 0.7, 0.9]$. It is clear that from Scenario 1 to Scenario 3, the level of data heterogeneity increases. 

Fig.~\ref{fig:impact_data_noniid} shows the accuracy lost for improving local and global fairness. The accuracy loss for reducing both local and global disparity from 0.3 to 0.05 in Scenario~1 is 0.045, which is 5.5\% of the initial accuracy. The accuracy loss is 7.9\% of the initial accuracy for Scenario~2 and 8.1\% for Scenario~3. The result indicate that the accuracy loss for achieving both fairness increases as the level of data heterogeneity increases.

When local disparity is bounded by 0.05, we observe that as the global disparity is reduced from 0.3 to 0.05, the change of accuracy is 0\%~\((0.736 \rightarrow 0.736)\) of the initial accuracy for Scenario~1, 0.4\%~\((0.715 \rightarrow 0.712)\) of the initial accuracy for Scenario~2, and 2.5\% ($0.719 \rightarrow 0.701$) for Scenario~3.  This suggests that in scenarios with higher data heterogeneity, the accuracy loss for achieving global fairness is higher.

 \textbf{Privacy Protection of the Local Statistics}
Local differential privacy (DP) mechanism can be applied when communicating local statistics in step 2 of algorithm in Section \ref{sec: training_steps} to protect client-level privacy.  Experimental results in Table \ref{Table_with_privacy}  shows how DP mechanisms affect the fairness and accuracy of our algorithms on UCI Adult. We add the Laplace mechanism on local statistics in Eq. \eqref{eq: est_statistics} so those local statistics satisfies $\epsilon$- differential privacy privacy, a smaller $
\epsilon$ indicate better privacy protection. The detail of Laplace mechanism is in Appendix \ref{app:pravicy}. The local disparity, global disparity and Avg-Acc after post-processing are reported in Table \ref{Table_with_privacy}


\begin{table}[h!] 
\centering
\fontsize{9pt}{9pt}\selectfont
\caption{Local Disparity (Local), Global Disparity (Global) and Avg-Acc under  $\epsilon$- Differential Privacy}
\label{Table_with_privacy}
\begin{tabular}{p{1.5 cm}|p{1.5cm}|p{1.5cm}|p{1.5 cm}} 
\toprule
\textbf{$\epsilon$}  & \textbf{Local} &  \textbf{Global } & \textbf{Avg-Acc} \\
\midrule
$\infty$ &  0.019  & 0.015& 81.30 \\
$1$  & 0.024 & 0.017 & 81.69 \\
$0.5$ &  0.043 & 0.038 & 81.32 \\
$0.1$ & 0.057 &0.072 & 82.00 \\
$0.05$ & 0.076 & 0.072 & 81.54 \\
\text{FedAvg} &  0.287 & 0.215 & 84.83 \\
\bottomrule
\end{tabular}
\end{table}

Compared to the initial FedAvg with $\epsilon = 0.05$, our framework reduces local disparity by 74\% (0.287 $\rightarrow$ 0.076) and global disparity  by 66\% (0.215 $\rightarrow$ 0.072). It demonstrates our framework's effectiveness in mitigating both local and global disparities under a $0.05$-differentially private setting. As $\epsilon$ decreases (i.e.,  better privacy protection), both local and global disparity increases. It is clear that there is a trade-off between privacy and fairness.

\section{Conclusions}

This paper presents a convex program to identify the optimal accuracy under varying levels of local and global fairness in multi-class FL. Our framework is implemented by solving an LP with polynomial time complexity. Experiments show it outperforms baselines in accuracy lost for fairness, communication efficiency, and computational cost for achieving fairness.

\section{Acknowledgments}

This work was supported by the National Science Foundation under Grant No. CNS-2228092.

\bibliographystyle{plainnat} 
\bibliography{refs} 

\newpage

\section*{Checklist}


 \begin{enumerate}

 \item For all models and algorithms presented, check if you include:
 \begin{enumerate}
   \item A clear description of the mathematical setting, assumptions, algorithm, and/or model. [Yes]
   
   The mathematical settings are in Section \ref{sec:Preliminary}
   \item An analysis of the properties and complexity (time, space, sample size) of any algorithm. [Yes]
   
   The time complexity of the implementation is analyzed in Section \ref{sec:lp}
   \item (Optional) Anonymized source code, with specification of all dependencies, including external libraries. [Yes]
   
   The code is included in supplementary of the paper submission. 
 \end{enumerate}

 \item For any theoretical claim, check if you include:
 \begin{enumerate}
   \item Statements of the full set of assumptions of all theoretical results. [Not Applicable]
   \item Complete proofs of all theoretical results. [Yes]

   All proofs are in Appendix.
   \item Clear explanations of any assumptions. [Not Applicable]     
 \end{enumerate}

 \item For all figures and tables that present empirical results, check if you include:
 \begin{enumerate}
   \item The code, data, and instructions needed to reproduce the main experimental results (either in the supplemental material or as a URL). [Yes]
   
   The code is included in supplementary of the paper submission. 
   \item All the training details (e.g., data splits, hyperparameters, how they were chosen). [Yes]
         \item A clear definition of the specific measure or statistics and error bars (e.g., with respect to the random seed after running experiments multiple times). [Yes]

         Those details are in Appendix
         \item A description of the computing infrastructure used. (e.g., type of GPUs, internal cluster, or cloud provider). [Yes]

         The GPU is talked in Experiments part of this paper.
 \end{enumerate}

 \item If you are using existing assets (e.g., code, data, models) or curating/releasing new assets, check if you include:
 \begin{enumerate}
   \item Citations of the creator If your work uses existing assets. [Yes]
   \item The license information of the assets, if applicable. [Not Applicable]
   \item New assets either in the supplemental material or as a URL, if applicable. [Not Applicable]
   \item Information about consent from data providers/curators. [Not Applicable]
   \item Discussion of sensible content if applicable, e.g., personally identifiable information or offensive content. [Not Applicable]
 \end{enumerate}

 \item If you used crowdsourcing or conducted research with human subjects, check if you include:
 \begin{enumerate}
   \item The full text of instructions given to participants and screenshots. [Not Applicable]
   \item Descriptions of potential participant risks, with links to Institutional Review Board (IRB) approvals if applicable. [Not Applicable]
   \item The estimated hourly wage paid to participants and the total amount spent on participant compensation. [Not Applicable]
 \end{enumerate}

 \end{enumerate}
\appendix
\onecolumn
\section*{\centering The Cost of Local and Global Fairness in Federated Learning: \\ Supplementary Materials}
\addcontentsline{toc}{section}{Appendix}

\section{The Parameters of Convex Program in Proposition 
\ref{prop:convex_program}}\label{app: parameters of LP}
This section provides the parameters of the convex program in Proposition~\ref{prop:convex_program}. 
We first define the following statistics for the client group \(D\):

\[
\begin{split}
    p_{ac}^y &= {\rm Pr}_D(Y = y, A = a, C = c) \\
    \alpha_{a}^{y} &= {\rm Pr}_D(Y = y, A = a)
\end{split}
\]

It is convenient to define the following matrix based on these statistics:

\begin{eqnarray*} \label{eq:parameters}
\mathbf{c}_{ac}^T &=& 
\left[ \begin{array}{cccc}
-p_{ac}^{1} & -p_{ac}^{2} & \cdots & -p_{ac}^{N} 
\end{array} \right] \in \mathbb{R}^{N} \\ [0.2cm]
\mathbf{n}_{ac}^{i} &=& \frac{1}{\alpha_a^i} \mathbf{E}_i \mathbf{c}_{ac}\in \mathbb{R}^N \\ [0.2cm]
\mathbf{m}^i &=& \mathbf{e}^i \in \mathbb{R}^N \\ [0.2cm]
\mathbf{N}_i &=& 
\left[ 
\begin{array}{cc}
(\mathbf{n}_{0i}^1)^T & -(\mathbf{n}_{1i}^1)^T \\ [0.2cm]
(\mathbf{n}_{0i}^2)^T & -(\mathbf{n}_{1i}^2)^T \\ 
\vdots & \vdots \\
(\mathbf{n}_{0i}^N)^T & -(\mathbf{n}_{1i}^N)^T 
\end{array} 
\right] \in \mathbb{R}^{N \times 2N} \\ [0.2cm]
\mathbf{M} &=& 
\left[ 
\begin{array}{cc}
(\mathbf{m}^1)^T & -(\mathbf{m}^1)^T \\ [0.2cm]
(\mathbf{m}^2)^T & -(\mathbf{m}^2)^T \\
\vdots & \vdots \\
(\mathbf{m}^N)^T & -(\mathbf{m}^N)^T 
\end{array} 
\right] \in \mathbb{R}^{N \times 2N}
\end{eqnarray*}

 where \(\mathbf{E}_i \in \mathbb{R}^{N \times N}\) is a matrix with the element at \((i, i)\) (row \(i\), column \(i\)) equal to 1 and all other elements are 0.  
\(\mathbf{e}^{i} \in \mathbb{R}^N\) is the elementary basis vector in Euclidean space, i.e., \(e^1 = [1, 0, 0, \dots, 0]^T\).

The variables of the convex program, which represent true positives of the outcome predictor  are:

\[
\mathbf{z}_{ac}^T = 
\left[ 
\begin{array}{cccc}
z_{ac}^{1} & z_{ac}^{2} & \cdots & z_{ac}^{N} 
\end{array} 
\right] \in \mathbb{R}^{N}
\]

The convex program (\ref{eq:convex_program}) is:

\begin{eqnarray}
\begin{array}{ll}
\text{minimize:} & \mathbf{c}^T \mathbf{z} \\
\text{with respect to:} & \mathbf{z} \in \mathbb{R}^{2NK} \\
\text{subject to:} & -\mathbf{b} \leq \mathbf{A} \mathbf{z} \leq \mathbf{b} \\
& \mathbf{z}_{ac} \in D_{ac}
\end{array}
\end{eqnarray}
where \(\mathbf{A}\), \(\mathbf{b}\), and \(\mathbf{c}\) are composed of matrices we defined above:

\[
\mathbf{c}^T = 
\left[ 
\begin{array}{ccccccc}
\mathbf{c}_{01}^T & \mathbf{c}_{11}^T & \mathbf{c}_{02}^T & \mathbf{c}_{12}^T & \cdots & \mathbf{c}_{0K}^T & \mathbf{c}_{1K}^T 
\end{array} 
\right] \in \mathbb{R}^{2NK}
\]

\[
\mathbf{z}^T = 
\left[ 
\begin{array}{ccccccc}
\mathbf{z}_{01}^T & \mathbf{z}_{11}^T & \mathbf{z}_{02}^T & \mathbf{z}_{12}^T & \cdots & \mathbf{z}_{0K}^T & \mathbf{z}_{1K}^T 
\end{array} 
\right] \in \mathbb{R}^{2NK}
\]

\renewcommand{\arraystretch}{1.5}
\[
\mathbf{A} = 
\left[ 
\begin{array}{cccc}
\mathbf{N}_1 & \mathbf{N}_2 & \cdots & \mathbf{N}_K \\
\mathbf{M} & \mathbf{0} & \cdots & \mathbf{0} \\
\mathbf{0} & \mathbf{M} & \cdots & \mathbf{0} \\
\vdots & \vdots & \ddots & \vdots \\
\mathbf{0} & \mathbf{0} & \cdots & \mathbf{M}
\end{array} 
\right] \in \mathbb{R}^{(N(K+1)) \times 2NK}
\]

\[
\mathbf{b}^T = 
\left[ 
\epsilon_0, \dots, \epsilon_0, \epsilon_1, \dots, \epsilon_1, \dots, \epsilon_K, \dots, \epsilon_K 
\right] \in \mathbb{R}^{N(K+1)}
\]

\section{The Parameters of Simplex $\widehat{D_{ac}}$}\label{app:convex_hull}
The N-dimensional polytope $\widehat{D_{ac}}$ can be defined using (N+1) inequalities. Any single point $\mathbf{u} \in \mathbb{R}^N$ lies in the $D_{ac}$ must have:
\begin{equation}
    \mathbf{K}_{ac}\mathbf{u}\leq \mathbf{l}_{ac}
\end{equation}
The true positives for predictor $\widetilde{Y}_{\mathbf{1}}$ that takes value of Eq. \ref{eq:optimal_predictor} are denoted as ${\rm TP}_{ac}^1 (\widetilde{Y}_{\boldsymbol{1}})$.

then,
{\small \[
\mathbf{K}_{ac}=
{\displaystyle \left[ \begin{array}{ccccc}
-1 & -1 & -1& \cdots &
-1\\
1-{\displaystyle \sum_{i\in \mathcal{Y}, i\neq 1}}{\rm TP}_{ac}^i (\widetilde{Y}_{\boldsymbol{1}})& {\rm TP}_{ac}^1 (\widetilde{Y}_{\boldsymbol{1}})& {\rm TP}_{ac}^1 (\widetilde{Y}_{\boldsymbol{1}})& \cdots & {\rm TP}_{ac}^1 (\widetilde{Y}_{\boldsymbol{1}})\\[0.15cm]
{\rm TP}_{ac}^2 (\widetilde{Y}_{\boldsymbol{1}})&1-{\displaystyle \sum_{i\in \mathcal{Y}, i\neq 2}}{{\rm TP}^{i}_{ac}}(\widetilde{Y}_{\mathbf{1}})&{\rm TP}_{ac}^2 (\widetilde{Y}_{\boldsymbol{1}})&\cdots &{\rm TP}_{ac}^2 (\widetilde{Y}_{\boldsymbol{1}})\\[0.15cm]
{\rm TP}_{ac}^3 (\widetilde{Y}_{\boldsymbol{1}})&{\rm TP}_{ac}^3 (\widetilde{Y}_{\boldsymbol{1}})&1-{\displaystyle \sum_{i\in \mathcal{Y}, i\neq 3}}{\rm TP}_{ac}^i (\widetilde{Y}_{\boldsymbol{1}})&\cdots &{\rm TP}_{ac}^3 (\widetilde{Y}_{\boldsymbol{1}})\\[0.15cm]
\vdots&\vdots&\vdots&\ddots&\vdots\\[0.15cm]
{\rm TP}_{ac}^N (\widetilde{Y}_{\boldsymbol{1}})&{\rm TP}_{ac}^N (\widetilde{Y}_{\boldsymbol{1}})&{\rm TP}_{ac}^N (\widetilde{Y}_{\boldsymbol{1}})&\cdots &1-{\displaystyle \sum_{i\in \mathcal{Y}, i\neq N}}{\rm TP}_{ac}^i (\widetilde{Y}_{\boldsymbol{1}})
\end{array} \right]}
\in \mathbb{R}^{(N+1)\times N}
\]}
\begin{eqnarray*}
\mathbf{l}_{ac} &=& \left[ \begin{array}{cccccc}
-1&{\rm TP}_{ac}^1 (\widetilde{Y}_{\boldsymbol{1}})& {\rm TP}_{ac}^2 (\widetilde{Y}_{\boldsymbol{1}})&{\rm TP}_{ac}^3 (\widetilde{Y}_{\boldsymbol{1}})& \cdots&{\rm TP}_{ac}^N (\widetilde{Y}_{\boldsymbol{1}})\end{array}\right]^T 
\in \mathbb{R}^{N+1}
\end{eqnarray*}

\section{The Parameters of LAE in Proposition \ref{prop:uniqueness} }
\label{app:para_LAE}
The true positives for predictor $\widetilde{Y}_{\mathbf{1}}$ that takes value of Eq. \ref{eq:optimal_predictor} are ${\rm TP}_{ac}^1 (\widetilde{Y}_{\boldsymbol{1}})$ and $\mathbf{z}_{ac}$ are the solution of the LP \eqref{eq: lp}. The parameters of LAE \eqref{eq:ugh-ugh}:  $ \mathbf{G}_{ac} \boldsymbol{\beta}_{ac}= \boldsymbol{\gamma}_{ac}$ are:

 {\small
    \begin{equation}
    \mathbf{G}_{ac}= {\displaystyle \left[ \begin{array}{cccccc}
    1 & 1 & 1& \cdots & 1&
    1\\[0.2cm]
    {\rm TP}_{ac}^1 (\widetilde{Y}_{\boldsymbol{1}}) & 1& 0& \cdots & 0 &0\\[0.2cm]
    {\rm TP}_{ac}^2 (\widetilde{Y}_{\boldsymbol{1}}) &0 &1 & \cdots &0 &0\\[0.2cm]
    \vdots& \vdots&  \vdots& \ddots& \vdots &\vdots\\[0.2cm]
    {\rm TP}_{ac}^{N-1} (\widetilde{Y}_{\boldsymbol{1}}) &0&0 &\cdots &1 &0\\[0.2cm]
    {\rm TP}_{ac}^N (\widetilde{Y}_{\boldsymbol{1}})&0&0&\cdots&0&1
    \end{array} \right]}, \quad
   \boldsymbol{\gamma}_{ac} = \left[\begin{array}{c}
    1  \\
    \mathbf{z}_{ac}
    \end{array} \right]
    \end{equation}
    }

\section{ Extension to Other Fairness Metrics (Equal Opportunity and Statistical Parity)}
\label{app:extension_to_other_fairness}
\subsection{Equal Opportunity}
Equal Opportunity is a special case of Equalized Odds. It requires equal true positive rates between the sensitive and non-sensitive for a given class $y$, typically $y=1$, rather than for all classes. The definitions of local and global Equal Opportunity are as follows:

\begin{definition}
Let $\epsilon_0 \in [0,1]$. The outcome predictor
$\widetilde{Y} : \mathcal{X} \times \mathcal{A} \times \mathcal{C} \rightarrow \mathcal{Y}$ for client group $D$ satisfies
\textbf{$\epsilon_0$-global equal opportunity} if
\begin{equation}
\begin{split}
&|{\rm Pr}_D \left\{ \widetilde{Y}(X,A,C) = 1\, | \, Y=1, A=1 \right\}  \\
-&{\rm Pr}_D \left\{ \widetilde{Y}(X,A,C) = 1 \, | \, Y=1, A=0 \right\}| \leq \epsilon_0.
\end{split}
\end{equation}
\end{definition}

\begin{definition}
($\boldsymbol{\epsilon_l}$-Local Fairness) Let $\boldsymbol{\epsilon}_l=[\epsilon_1, \cdots, \epsilon_K] \in [0,1]^K$. The outcome predictor
$\widetilde{Y} : \mathcal{X} \times \mathcal{A} \times \mathcal{C} \rightarrow \mathcal{Y}$ satisfies
\textbf{$\boldsymbol{\epsilon}_l$-local equal opportunity} if
\begin{equation}
\begin{split}
&|{\rm Pr}_D \left\{ \widetilde{Y}(X,A,C) = 1\, | \, Y=1, A=1, C=c \right\}\\ - 
&{\rm Pr}_D \left\{ \widetilde{Y}(X,A,C) = 1 \, | \, Y=1, A=0, C=c \right\}| \leq \epsilon_c.
\end{split}
\end{equation}
\end{definition}

The LP for Equal Opportunity is a special case of LP \eqref{eq: lp}, where only the constraints for class $y=1$ are retained. The LP for Equal Opportunity is:

\begin{eqnarray}
\begin{array}{ll}
\mbox{minimize:} & \mathbf{c}^T \mathbf{z} \\
\mbox{with respect to:} & \mathbf{z} \in \mathbb{R}^{2NK} \\
\mbox{subject to:} & -\mathbf{b}\leq \mathbf{Az} \leq \mathbf{b} \\
& \mathbf{K}_{ac}\mathbf{z}_{ac} \leq \mathbf{l}_{ac}
\end{array}
\end{eqnarray}
with
\begin{eqnarray*}
\mathbf{c}^T &=& \left[ \begin{array}{ccccccc}
\mathbf{c}_{01}^T & \mathbf{c}_{11}^T & \mathbf{c}_{02}^T & \mathbf{c}_{12}^T & \cdots & \mathbf{c}_{0K}^T & \mathbf{c}_{1K}^T \end{array} \right] \in \mathbb{R}^{2NK}\\
\mathbf{z}^T &=& \left[ \begin{array}{ccccccc}
\mathbf{z}_{01}^T & \mathbf{z}_{11}^T & \mathbf{z}_{02}^T& \mathbf{z}_{12}^T & \cdots & \mathbf{z}^T_{0K}& \mathbf{z}^T_{1K} \end{array} \right] \in \mathbb{R}^{2NK}
\\
\end{eqnarray*}
\renewcommand{\arraystretch}{1.5} 
\[
\mathbf{A} = {\displaystyle \left[ \begin{array}{cccc}
\mathbf{N}_1& \mathbf{N}_2 & \cdots & \mathbf{N}_{K}\\
\mathbf{M} & \mathbf{0}& \cdots& \mathbf{0}\\
\mathbf{0} & \mathbf{M}& \cdots& \mathbf{0}\\
\vdots & \vdots&  \ddots &\vdots\\
\mathbf{0} & \mathbf{0}& \cdots& \mathbf{M}\\
\end{array} \right]} \in \mathbb{R}^{(K+1)\times 2NK}
\]
\begin{eqnarray*}
\mathbf{b}^T &=&[\epsilon_0, \epsilon_1,\cdots, \epsilon_K]  \in \mathbb{R}^{(K+1)}
\end{eqnarray*}

The parameters are the same as those for Equalized Odds in Appendix~\ref{app: parameters of LP}, except for Equal Opportunity, where \(\mathbf{N}_i\) and \(\mathbf{M}\) are defined as:

\begin{eqnarray*}
\mathbf{N}_i &=& 
\left[\begin{array}{cc}
(\mathbf{n}_{0i}^1)^T & -(\mathbf{n}_{1i}^1)^T 
\end{array}\right] \in \mathbb{R}^{1 \times 2N} \\[0.2cm]
\mathbf{M} &=& 
\left[\begin{array}{cc}
(\mathbf{m}^1)^T & -(\mathbf{m}^1)^T 
\end{array}\right] \in \mathbb{R}^{1 \times 2N}
\end{eqnarray*}

$\mathbf{K}_{ac}, \mathbf{l}_{ac}$ is the same as the LP \eqref{eq: lp} for Equalized Odds, which are in Appendix \ref{app:convex_hull}.

\subsection{Statistical Parity}
\begin{definition}
  Let $\epsilon_0 \in [0,1]$, the outcome predictor
$\widetilde{Y} : \mathcal{X} \times \mathcal{A} \times \mathcal{C} \rightarrow \mathcal{Y}$ for client group $D$ satisfies
\textbf{$\epsilon_0$-global statistical parity} if for all $y \in \mathcal{Y}$,  
\begin{equation}
\begin{split}
&|{\rm Pr}_D \left\{ \widetilde{Y}(X,A,C) = y\, | \ A=1 \right\}  \\
-&{\rm Pr}_D \left\{ \widetilde{Y}(X,A,C) = y \, | \ A=0 \right\}| \leq \epsilon_0.
\end{split}
\end{equation}
\end{definition}

\begin{definition}
($\boldsymbol{\epsilon_l}$-Local Fairness) Let $\boldsymbol{\epsilon}_l=[\epsilon_1 \cdots, \epsilon_K] \in [0,1]^K$ , the outcome predictor
$\widetilde{Y} : \mathcal{X} \times \mathcal{A} \times \mathcal{C} \rightarrow \mathcal{Y}$ satisfies
\textbf{$\boldsymbol{\epsilon}_l$-local statistical parity} if
\begin{equation}
\begin{split}
&|{\rm Pr}_D \left\{ \widetilde{Y}(X,A,C) = y\, | \, Y=y, A=1, C=c \right\}\\ - 
&{\rm Pr}_D \left\{ \widetilde{Y}(X,A,C) = y \, | \, Y=y, A=0, C=c \right\}| \leq \epsilon_c.
\end{split}
\end{equation}
\end{definition}

The variables we use to characterize the LP for statistical parity differ from those used for Equalized Odds and Equal Opportunity. Statistical Parity requires that the predictor's positive rate be the same for both sensitive and non-sensitive groups. Therefore, we need to focus on both true positives and false positives for a given class.

In the statistical parity setting, the variables we use to characterize the outcome predictor $\widetilde{Y}: \mathcal{X}\times \mathcal{A} \times \mathcal{C} \rightarrow \mathcal{Y}$ are:

\begin{equation}
    z_{ac}^{yj}= {\rm Pr}(\widetilde{Y}=y| \widetilde{Y}_{\mathbf{1}}=j, A=a, C=c)
\end{equation}

The value of $\widetilde{Y}$ only depends on the output of the predictor $\widetilde{Y}_{\mathbf{1}}$ that is defined in \eqref{eq:optimal_predictor}, client label $c$ and group $a$.

We find it is convenient to  define the following probability of client group $D$ for formulating the LP:
\begin{equation*}
\begin{split}
    u_{ac}^{yj}= &{\rm Pr}_D (Y=y, \widetilde{Y}_{\mathbf{1}}=j, A=a,C=c)\\
    u_{ac}^j= & {\rm Pr}_D (\widetilde{Y}_{\mathbf{1}}=j, A=a, C=c)\\
    u_{ac}=& {\rm Pr}_D(A=a, C=c)\\
    u_a =& {\rm Pr}_D(A=a)
\end{split}
\end{equation*}
The LP for statistical Parity is:
\begin{eqnarray} \label{eq:LP_SP}
\begin{array}{ll}
\mbox{maximize:} & {\displaystyle \sum_{c \in \mathcal{C}} \sum_{a \in \mathcal{A}} \sum_{j \in \mathcal{Y}} \sum_{y \in \mathcal{Y}}} u_{ac}^{yj}z_{ac}^{yj} \\
\mbox{with respect to:}& z_{ac}^{yj} \in \mathbb{R}, \forall y, j \in \mathcal{Y}, a \in \mathcal{A}, c \in \mathcal{C} \\
\mbox{subject to:} & \forall y \in \mathcal{Y}, -\epsilon_0 \leq {\displaystyle \sum_{c\in \mathcal{C}} \sum_{j \in \mathcal{Y}} \frac{z_{0c}^{yj} \cdot u_{0c}^j}{u_0}}- {\displaystyle \sum_{c\in \mathcal{C}} \sum_{j \in \mathcal{Y}} \frac{z_{1c}^{yj} \cdot u_{1c}^j}{u_1}} \leq \epsilon_0\\
& \forall y \in \mathcal{Y}, c \in \mathcal{C},  -\epsilon_c \leq {\displaystyle  \sum_{j \in \mathcal{Y}} \frac{z_{0c}^{yj} \cdot u_{0c}^j}{u_{0c}}}- {\displaystyle  \sum_{j \in \mathcal{Y}} \frac{z_{1c}^{yj} \cdot u_{1c}^j}{u_{1c}}} \leq \epsilon_c\\
& \forall c \in \mathcal{C}, a \in \mathcal{A}, j \in \mathcal{Y}, {\displaystyle \sum_{y \in \mathcal{Y}}} z_{ac}^{yj}=1\\
& \forall y,j \in \mathcal{Y}, a \in \mathcal{A}, c \in \mathcal{C}, 0 \leq z_{ac}^{yj} \leq 1
\end{array}
\end{eqnarray}

Then, the outcome predictor $\widetilde{Y}: \mathcal{X} \times \mathcal{A} \times \mathcal{C} \rightarrow \mathcal{Y}$ that takes values:
\begin{equation}\label{eq:predictor_SP}
     \widetilde{Y}_{\mathbf{1}}=j, A=a, C=c: \widetilde{Y}=y \text{ with probability } z_{ac}^{yj}
\end{equation}
satisfies local and global Statistical Parity. The proof of this assertion is given below.

\textbf{Proof}: We first show that the outcome predictor that takes value of \eqref{eq:predictor_SP} satisfies global statistical parity:

The first constraint of LP \eqref{eq:LP_SP} is, $\forall y \in \mathcal{Y}$:
\begin{equation}
    \begin{split}
        -\epsilon_0 \leq & {\displaystyle \sum_{c\in \mathcal{C}} \sum_{j \in \mathcal{Y}} \frac{z_{0c}^{yj} \cdot u_{0c}^j}{u_0}} - {\displaystyle \sum_{c\in \mathcal{C}} \sum_{j \in \mathcal{Y}} \frac{z_{1c}^{yj} \cdot u_{1c}^j}{u_1}} \leq \epsilon_0 \\
        \iff -\epsilon_0 \leq & {\displaystyle \sum_{c \in \mathcal{C}} \sum_{j \in \mathcal{Y}}\frac{{\rm Pr}_D(\widetilde{Y}=y| \widetilde{Y}_{\mathbf{1}}=j,  C=c, A=0)\cdot {\rm Pr}_D(\widetilde{Y}_{\mathbf{1}}=j, A=0, C=c)}{{\rm Pr}_D(A=0)}} \\
        & - {\displaystyle \sum_{c \in \mathcal{C}} \sum_{j \in \mathcal{Y}}\frac{{\rm Pr}_D(\widetilde{Y}=y| \widetilde{Y}_{\mathbf{1}}=j, C=c, A=1)\cdot {\rm Pr}_D(\widetilde{Y}_{\mathbf{1}}=j, A=1, C=c)}{{\rm Pr}_D(A=1)}} \leq \epsilon_0\\
        \iff -\epsilon_0 \leq& \frac{{\rm Pr}_D( \widetilde{Y}=y, A=0)}{{\rm Pr}_D (A=0)} - \frac{{\rm Pr}_D( \widetilde{Y}=y, A=1)}{{\rm Pr}_D (A=1)} \leq \epsilon_0\\
        \iff -\epsilon_0 \leq & {\rm Pr}_D (\widetilde{Y}=y|A=0) - {\rm Pr}_D (\widetilde{Y}=y |A=1) \leq \epsilon_0
    \end{split}
\end{equation}
The first constraint of the LP is equivalent to the global statistical parity. The outcome predictor that satisfies the LP will satisfies the global $\epsilon_0-$ statistical parity.

Then, we show that the outcome predictor \eqref{eq:predictor_SP} satisfies local statistical parity:

The second constraint of the LP \eqref{eq:LP_SP} is, $\forall y \in \mathcal{Y}, c\in \mathcal{C}$:
\begin{equation}
    \begin{split}
        -\epsilon_c \leq & {\displaystyle  \sum_{j \in \mathcal{Y}} \frac{z_{0c}^{yj} \cdot u_{0c}^j}{u_{0c}}} - {\displaystyle  \sum_{j \in \mathcal{Y}} \frac{z_{1c}^{yj} \cdot u_{1c}^j}{u_{1c}}} \leq \epsilon_c \\
        \iff -\epsilon_c \leq & {\displaystyle  \sum_{j \in \mathcal{Y}} \frac{{\rm Pr}_D(\widetilde{Y}=y| \widetilde{Y}_{\mathbf{1}}=j,  C=c, A=0)\cdot {\rm Pr}_D(\widetilde{Y}_{\mathbf{1}}=j, A=0, C=c)}{{\rm Pr}_D(A=0, C=c)}} \\
        & - {\displaystyle  \sum_{j \in \mathcal{Y}} \frac{{\rm Pr}_D(\widetilde{Y}=y| \widetilde{Y}_{\mathbf{1}}=j, C=c, A=1)\cdot {\rm Pr}_D(\widetilde{Y}_{\mathbf{1}}=j, A=1, C=c)}{{\rm Pr}_D(A=1, C=c)}} \leq \epsilon_c \\
        \iff -\epsilon_c \leq & \frac{{\rm Pr}_D( \widetilde{Y}=y, A=0, C=c)}{{\rm Pr}_D(A=0, C=c)} - \frac{{\rm Pr}_D( \widetilde{Y}=y, A=1, C=c)}{{\rm Pr}_D(A=1, C=c)} \leq \epsilon_c\\
        \iff -\epsilon_c \leq & {\rm Pr}_D(\widetilde{Y}=y|A=0, C=c) - {\rm Pr}_D (\widetilde{Y}=y|A=1, C=c) \leq \epsilon_c
    \end{split}
\end{equation}

The second constraints of the LP is equivalent to local statistical parity, thus, it will lead a outcome predictor that satisfies $\boldsymbol{\epsilon}_l-$ local statistical parity. 

Since the variables $z_{ac}^{yj}$ represent the probability, it satisfies the third and fourth constraints. 

The objective function we maximize is: 
\begin{equation}
    \begin{split}
        &{\displaystyle \sum_{c \in \mathcal{C}} \sum_{a \in \mathcal{A}} \sum_{j \in \mathcal{Y}} \sum_{y \in \mathcal{Y}}} u_{ac}^{yj}z_{ac}^{yj}\\
        =& {\displaystyle \sum_{c \in \mathcal{C}} \sum_{a \in \mathcal{A}} \sum_{j \in \mathcal{Y}} \sum_{y \in \mathcal{Y}}} {\rm Pr}(\widetilde{Y}=y| \widetilde{Y}_{\mathbf{1}}=j, A=a, C=c) \cdot {\rm Pr}_D( Y=y, \widetilde{Y}_{\mathbf{1}}=j, A=a, C=c)\\
        =&  {\displaystyle \sum_{c \in \mathcal{C}} \sum_{a \in \mathcal{A}} \sum_{j \in \mathcal{Y}} \sum_{y \in \mathcal{Y}}} {\rm Pr}(\widetilde{Y}=y| \widetilde{Y}_{\mathbf{1}}=j, A=a, C=c, Y=y) \cdot {\rm Pr}_D( Y=y, \widetilde{Y}_{\mathbf{1}}=j, A=a, C=c) \\
        =& {\displaystyle \sum_{c \in \mathcal{C}} \sum_{a \in \mathcal{A}} \sum_{j \in \mathcal{Y}} \sum_{y \in \mathcal{Y}}} {\rm Pr}(\widetilde{Y}=y,  \widetilde{Y}_{\mathbf{1}}=j, A=a, C=c, Y=y)\\
        =& {\displaystyle  \sum_{y \in \mathcal{Y}}} {\rm Pr}(\widetilde{Y}=y, Y=y)\\
        =& {\rm Pr}_D (\widetilde{Y}=Y)
    \end{split}
\end{equation}
The second equation holds because $\widetilde{Y}$ depends only on $\widetilde{Y}_{\mathbf{1}}, A$, and $C$. $\widetilde{Y}$ is independent of $Y$ given $\widetilde{Y}_{\mathbf{1}}, A$, and $C$. The objective we maximize represents the accuracy of the outcome predictor.

Thus, the outcome predictor in \eqref{eq:predictor_SP} satisfies local and global fairness and maximizes the accuracy under our LP.
\section{Theoretical Proofs}
\subsection{Proof of Proposition \ref{prop:NP_multi_class}}
\label{app:proof_NP_multi_class}
\textbf{Proposition \ref{prop:NP_multi_class}}     Consider client $c$ and group $a$, let $\{\widetilde{Y}_{\boldsymbol{\theta}}\}_{\boldsymbol{\theta} \in \mathbb{R}^N_{\geq 0}}$ be the set of all derived outcome predictors, and let $\{\phi_g\}_{g=1}^{N-1}, \in [0,1]^N$ be a set of specified values of true positives. If $\widetilde{Y}^*: \mathcal{X}\times \mathcal{A} \times \mathcal{C} \rightarrow \mathcal{Y}$ is the solution to the following optimization problem:
    \begin{eqnarray} \label{eq:NP_optimization_problem}
    \begin{array}{ll}
    \text{maximize} & {\rm TP}_{ac}^N (\widetilde{Y})\\
    \text{with respect to} & \widetilde{Y}: \mathcal{X} \times \mathcal{A} \times \mathcal{C} \rightarrow \mathcal{Y} \\
    \text{subject to } & \forall g \in \{1, 2, \cdots, N-1\}, \\
    &{\rm TP}_{ac}^g (\widetilde{Y}) = \phi_g
    \end{array}
    \end{eqnarray}
    then $\widetilde{Y}^* \in \{\widetilde{Y}_{\boldsymbol{\theta}}\}_{\boldsymbol{\theta} \in \mathbb{R}^N_{\geq 0}}$.
    
\textbf{Proof:} Our proposition is a special case of the NP multi-class classification problem defined in \citep{tian2024neyman}, where the number of constraints on true positives is chosen to be an arbitrary value less than \(N\), and the objective is to maximize the weighted sum of the remaining true positives. In our case, the number of constraints is fixed at \(N - 1\), and we maximize the true positive rate for class \(N\). 
Our proof follows the proof provided in \citep{tian2024neyman}.

We first consider the Lagrange function associated with problem \eqref{eq:NP_optimization_problem}:

\[
    L (\widetilde{Y}, \boldsymbol{\lambda}) = {\rm TP}_{ac}^N (\widetilde{Y}) + \sum_{g=1}^{N-1} \lambda_g \left( {\rm TP}_{ac}^g (\widetilde{Y}) - \phi_g \right)
\]
where $\boldsymbol{\lambda} = [\lambda_1, \lambda_2, \cdots, \lambda_{N-1}] \in \mathbb{R}^{N-1}$, with $\lambda_g$ as the Lagrange multiplier corresponding to the $g$-th constraint: ${\rm TP}_{ac}^g (\widetilde{Y}) = \phi_g$.

The dual problem of \eqref{eq:NP_optimization_problem} can be written as:
\begin{equation}\label{eq:dual_problem}
\max_{\boldsymbol{\lambda} \in \mathbb{R}^{N-1}} \min_{\widetilde{Y}} L(\widetilde{Y}, \boldsymbol{\lambda})
\end{equation}

Problem \eqref{eq:dual_problem} can be solved analytically. We consider $\widetilde{Y}^{\star}_{\boldsymbol{\lambda}}: \mathcal{X} \times \mathcal{A} \times \mathcal{C} \rightarrow \mathcal{Y}$ that minimizes the Lagrange function $L(\widetilde{Y}, \boldsymbol{\lambda})$ for any $\boldsymbol{\lambda} \in \mathbb{R}^{N-1}$.  
We find it convenient to represent the following probability as:
\[
    s_{ac}^y = {\rm Pr}_D(Y = y \mid A = a, C = c)
\]
The term ${\rm TP}_{ac}^y (\widetilde{Y})$ is equal to:
\begin{equation}\label{eq:eq_true_positive}
\begin{split}
    {\rm TP}_{ac}^y (\widetilde{Y}) &= \mathbb{E}_{{\rm Pr}_{X \mid Y, A, C}}[\mathbf{1}(\widetilde{Y} = y)] \\
    &= \mathbb{E}_{{\rm Pr}_{X \mid A, C}}[\mathbf{1}(\widetilde{Y} = y) \cdot r_y(X, a, c)] \cdot \frac{{\rm Pr}_D(A = a, C = c)}{{\rm Pr}_D(Y = y, A = a, C = c)} \\
    &= \mathbb{E}_{{\rm Pr}_{X \mid A, C}}[\mathbf{1}(\widetilde{Y} = y) \cdot r_y(X, a, c)] \cdot \frac{1}{{\rm Pr}_D(Y = y \mid A = a, C = c)} \\
    &= \mathbb{E}_{{\rm Pr}_{X \mid A, C}}[\mathbf{1}(\widetilde{Y} = y) \cdot r_y(X, a, c)] \cdot \frac{1}{s_{ac}^y}
\end{split}
\end{equation}
where $r_y(x, a, c)$ is the $y$-th element of the Bayesian optimal score function in Definition~4.1. Substituting Eq.~\eqref{eq:eq_true_positive} into the Lagrange function, we can express it as:
\[
\begin{split}
    L(\widetilde{Y}, \boldsymbol{\lambda}) =& \sum_{g=1}^{N-1} \frac{\lambda_g}{s_{ac}^g} \mathbb{E}_{{\rm Pr}_{X \mid A, C}}[\mathbf{1}(\widetilde{Y} = g) \cdot r_g(X, a, c)] \\
    &+ \mathbb{E}_{{\rm Pr}_{X \mid A, C}}[\mathbf{1}(\widetilde{Y} = N) \cdot r_N(X, a, c)] - \sum_{g=1}^{N-1} \lambda_g \phi_g
\end{split}
\]

For any given $\boldsymbol{\lambda}$, the predictor $\widetilde{Y}^*_{\boldsymbol{\lambda}}$ that minimizes the Lagrange function $L(\widetilde{Y}, \boldsymbol{\lambda})$ takes the values:
\begin{equation}\label{eq:app_y_optimal}
\begin{split}
    \widetilde{Y}^*_{\boldsymbol{\lambda}} &= g, \quad \text{if } \frac{\lambda_g}{s_{ac}^g} r_g(x, a, c) = \max \left\{ \frac{\lambda_1}{s_{ac}^1} r_1(x, a, c), \ldots, r_N(x, a, c) \right\} \\
    \text{or, }\widetilde{Y}^*_{\boldsymbol{\lambda}} &= N, \quad \text{if } r_N(x, a, c) = \max \left\{ \frac{\lambda_1}{s_{ac}^1} r_1(x, a, c), \ldots, r_N(x, a, c) \right\}
\end{split}
\end{equation}

We then consider the optimal $\boldsymbol{\lambda}^*$ that maximizes $L(\widetilde{Y}^*_{\boldsymbol{\lambda}}, \boldsymbol{\lambda})$. The first-order optimality conditions are:
\begin{equation}\label{eq:first_order_condition}
\begin{split}
    \forall g \in \{1, 2, \ldots, N-1\}, \quad &\frac{\partial L(\widetilde{Y}^*_{\boldsymbol{\lambda}}, \boldsymbol{\lambda})}{\partial \lambda_g} \bigg|_{\boldsymbol{\lambda}^*} = 0 \\
    \Longleftrightarrow & \quad {\rm TP}_{ac}^g (\widetilde{Y}_{\boldsymbol{\lambda}^*}) = \phi_g
\end{split}
\end{equation}

Consider the Bayesian optimal score function $r_y(X,A,C)$ with continuous probability density. Then there  exist $\boldsymbol{\lambda}^*$ that satisfies Eq.~\eqref{eq:first_order_condition}. The solution must satisfy:
\[
    \frac{\lambda_g^*}{s_{ac}^g} \geq 0, \quad \forall g \in \{1, \ldots, N-1\}
\]

If $\frac{\lambda_g}{s_{ac}^g} < 0$, the predictor will always output $N$, violating constraints when $\phi_g > 0$.

Under strong duality \citep{tian2024neyman}, the solution of the dual problem \eqref{eq:dual_problem} is equivalent to the solution of the primal problem \eqref{eq:NP_optimization_problem}. 
Based on \citep{tian2024neyman}, 
 in this problem, strong duality holds. 
Therefore, the solution of problem \eqref{eq:NP_optimization_problem} is $\widetilde{Y}^{\star}_{\boldsymbol{\lambda}^*}$, which is a derived outcome predictor as defined in Definition~\ref{def:derived_outcome_preditor}, with:
\[
    \theta_g = \frac{\lambda_g^*}{s_{ac}^g}, \quad \forall g \in \{1, \ldots, N-1\}, \quad \text{and } \theta_N = 1.
\]

\subsection{Proof of Proposition \ref{prop:convex_property}}
\label{app:proof_convex_property}
\textbf{Proposition \ref{prop:convex_property}}: Let $D_{ac}$ be the region defined in Def. \ref{region_under_ROC}. Then, $D_{ac}$ is a convex set.

 \textbf{Proof:} 
Consider any $\mathbf{r}_0, \mathbf{r}_1 \in [0,1]^N$, from Def. \ref{region_under_ROC}, if $\mathbf{r}_0, \mathbf{r}_1 \in D_{ac}$, then:
\begin{equation} \label{eq:26}
\begin{split}
    \mathbf{v}_{\boldsymbol{\theta}}^T \mathbf{r}_0 \leq \mathbf{v}_{\boldsymbol{\theta}}^T \textbf{TP}_{ac}(\widetilde{Y}_{\boldsymbol{\theta}}), \quad \forall \boldsymbol{\theta} \in \mathbb{R}^N_{\geq 0}, \\
    \mathbf{v}_{\boldsymbol{\theta}}^T \mathbf{r}_1 \leq \mathbf{v}_{\boldsymbol{\theta}}^T \textbf{TP}_{ac}(\widetilde{Y}_{\boldsymbol{\theta}}), \quad  \forall \boldsymbol{\theta} \in \mathbb{R}^N_{\geq 0}.
\end{split}
\end{equation}
 Since $\lambda \in [0,1]$, we must have:
\begin{equation} 
\begin{split}
    \mathbf{v}_{\boldsymbol{\theta}}^T (\lambda \mathbf{r}_0) & \leq \mathbf{v}_{\boldsymbol{\theta}}^T (\lambda \textbf{TP}_{ac}(\widetilde{Y}_{\boldsymbol{\theta}})), \quad \forall \boldsymbol{\theta} \in \mathbb{R}^N_{\geq 0}, \\
    \mathbf{v}_{\boldsymbol{\theta}}^T ((1-\lambda)\mathbf{r}_1) & \leq \mathbf{v}_{\boldsymbol{\theta}}^T ((1-\lambda) \textbf{TP}_{ac}(\widetilde{Y}_{\boldsymbol{\theta}})), \quad \forall \boldsymbol{\theta} \in \mathbb{R}^N_{\geq 0}, \\
    \Rightarrow \mathbf{v}_{\boldsymbol{\theta}}^T(\lambda \mathbf{r}_0 + (1-\lambda) \mathbf{r}_1) & \leq \mathbf{v}_{\boldsymbol{\theta}}^T \textbf{TP}_{ac}(\widetilde{Y}_{\boldsymbol{\theta}}), \quad \forall \boldsymbol{\theta} \in \mathbb{R}^N_{\geq 0}, \\
\end{split}
\end{equation}

which implies: $ \lambda \mathbf{r}_0 + (1-\lambda) \mathbf{r}_1 \in D_{ac}.$

We choose $\mathbf{r}_0, \mathbf{r}_1 \in D_{ac}$ arbitrarily. Thus, for all $\mathbf{r}_0, \mathbf{r}_1 \in D_{ac}$ and $\lambda \in [0,1]$, it holds that $\lambda \mathbf{r}_0 + (1-\lambda) \mathbf{r}_1 \in D_{ac}$. Therefore, $D_{ac}$ is a convex set.

\subsection{Proof of Proposition \ref{prop:true_positive_feasible}}
\label{app: proof_true_positive_feasible}
\textbf{Proposition \ref{prop:true_positive_feasible}:}Let $D_{ac}$ be the region as defined in Def.\ref{region_under_ROC}. For any predictor $\widetilde{Y}: \mathcal{X} \times \mathcal{A} \times \mathcal{C} \rightarrow \mathcal{Y}$, let the point representing true positives of $\widetilde{Y}$ be:  
$\textbf{TP}_{ac}(\widetilde{Y}) = [{\rm TP}^{y}_{ac}(\widetilde{Y}), \cdots, {\rm TP}^{N}_{ac}(\widetilde{Y})]^T$.
 Then, $ \textbf{TP}_{ac}(\widetilde{Y})$ lies in $D_{ac}$.

\textbf{Proof}: Proposition \ref{prop:true_positive_feasible} states that for any predictor $\widetilde{Y}: \mathcal{X} \times \mathcal{A} \times \mathcal{C} \rightarrow \mathcal{Y}$, the following condition must hold:

\begin{equation}
\begin{split}
    \textbf{TP}_{ac}(\widetilde{Y}) \in D_{ac}
    \iff \textbf{TP}_{ac}(\widetilde{Y}) &\in \underset{\boldsymbol{\theta}\in \mathbb{R}^N _{\geq 0}}{\bigcap} 
        \left\{\mathbf{x} \in [0,1]^N|\mathbf{v}_{\boldsymbol{\theta}}^T \mathbf{x}\leq \mathbf{v}_{\boldsymbol{\theta}}^T \textbf{TP}_{ac}(\widetilde{Y}_{\boldsymbol{\theta}}) \right\}\\
        \iff
        \forall \boldsymbol{\theta} \in
        \mathbb{R}^N_{\geq 0},&  \quad \mathbf{v}_{\boldsymbol{\theta}}^T \textbf{TP}_{ac}(\widetilde{Y}) \leq \mathbf{v}_{\boldsymbol{\theta}}^T \textbf{TP}_{ac}(\widetilde{Y}_{\boldsymbol{\theta}})
\end{split}
\end{equation}

To prove this, we consider the value of $\mathbf{v}_{\boldsymbol{\theta}}^T \textbf{TP}_{ac}(\widetilde{Y})$:

\begin{equation} \label{eq:vTtp}
\begin{split}
\mathbf{v}_{\boldsymbol{\theta}}^T \textbf{TP}_{ac}(\widetilde{Y}) &= \sum_{y \in \mathcal{Y}} \theta_y {\rm Pr}_D(Y=y \mid A=a, C=c) {\rm TP}_{ac}^y(\widetilde{Y})\\
&= \sum_{y \in \mathcal{Y}} \theta_y {\rm Pr}_D(Y=y \mid A=a, C=c) \mathbb{E}_{{\rm Pr}_{X \mid A,C}}[\mathbf{1}(\widetilde{Y} = y) \cdot r_y(X,a,c)] \cdot \frac{1}{{\rm Pr}_D(Y=y \mid A=a,C=c)} \\
&= \sum_{y \in \mathcal{Y}} \theta_y \mathbb{E}_{{\rm Pr}_{X \mid A,C}}[\mathbf{1}(\widetilde{Y} = y) \cdot r_y(X,a,c)] \\
&= \sum_{y \in \mathcal{Y}} \mathbb{E}_{{\rm Pr}_{X \mid A,C}}[\mathbf{1}(\widetilde{Y} = y) \cdot \theta_y r_y(X,a,c)]
\end{split}
\end{equation}

The second equation is from Eq. \eqref{eq:eq_true_positive} in Appendix \ref{app:proof_NP_multi_class}. Equation \eqref{eq:vTtp} achieves the  maximum value if:

\begin{equation} \label{eq:41}
\widetilde{Y} = y, \quad \text{if } \theta_y r_y(x,a,c) = \max_{i=1}^{N} \theta_i r_i(x,a,c)
\end{equation}

The predictor that takes the value of Eq. \ref{eq:41} is a derived outcome predictor $\widetilde{Y}_{\boldsymbol{\theta}}$ as defined in Def. \ref{def:derived_outcome_preditor}. The derived outcome predictor $\widetilde{Y}_{\boldsymbol{\theta}}$ maximizes the value of $\mathbf{v}_{\boldsymbol{\theta}}^T \textbf{TP}_{ac}(\widetilde{Y})$ for all $\boldsymbol{\theta} \in \mathbb{R}^N_{\geq 0}$. Therefore, any predictor must satisfy $\mathbf{v}_{\boldsymbol{\theta}}^T \textbf{TP}_{ac}(\widetilde{Y}) \leq \mathbf{v}_{\boldsymbol{\theta}}^T \textbf{TP}_{ac}(\widetilde{Y}_{\boldsymbol{\theta}})$, which is equivalent to $\textbf{TP}_{ac}(\widetilde{Y}) \in D_{ac}$.

\subsection{Proof of Proposition \ref{prop:convex_program}}
\label{app:proof_convex_program}
\textbf{Proposition \ref{prop:convex_program}: } Let the vector $\mathbf{z} \in \mathbb{R}^{2 N  K}$:
\begin{eqnarray*}
\mathbf{z}^T &=& \left[ \begin{array}{ccccccc}
\mathbf{z}_{01}^T & \mathbf{z}_{11}^T & \mathbf{z}_{02}^T& \mathbf{z}_{12}^T \cdots & \mathbf{z}^T_{0K}& \mathbf{z}^T_{1K} \end{array} \right]
\end{eqnarray*}
with 
\begin{eqnarray*}
    \mathbf{z}_{ac}^T &=& \left[ \begin{array}{cccccc}
  z_{ac}^{1}& z_{ac}^{2}& z_{ac}^{3}& \cdots &z_{ac}^{N} \end{array}\right] \in \mathbb{R}^N
\end{eqnarray*}
satisfy the following convex program
\begin{eqnarray}\label{eq:29}
\begin{array}{ll}
\mbox{minimize:} &\mathbf{c}^T \mathbf{z} \\
\mbox{with respect to:}& \mathbf{z}  \in \mathbb{R}^{2NK} \\
\mbox{subject to:} & -\mathbf{b}\leq \mathbf{Az} \leq \mathbf{b} \\
& \mathbf{z}_{ac} \in D_{ac}, \forall a \in \mathcal{A}, c \in \mathcal{C}
\end{array}
\end{eqnarray}
then, the outcome predictor $\widetilde{Y}:\mathcal{X} \times \mathcal{A} \times \mathcal{C} \rightarrow \mathcal{Y}$ that satisfies eq. \eqref{eq:3} for all $ y \in \mathcal{Y}, a\in \mathcal{A}, c\in \mathcal{C}$
\begin{equation} \label{eq:3}
     {\rm Pr}(\widetilde{Y}=y|Y=y, A=a,C=c)=z_{ac}^{y}
\end{equation}
is a $\boldsymbol{\epsilon}$-fair optimal outcome predictor.The optimal accuracy for a $\epsilon$-fair outcome predictor is $-\mathbf{c}^T\mathbf{z}.$

\textbf{Proof:} We first show that the outcome predictor $\widetilde{Y}$ in Proposition \ref{prop:convex_program} satisfies global Equalized Odds.

The probability ${\rm Pr}_D(\widetilde{Y}=y|Y=y, A=a)$ can be extended as follows:
\begin{equation}
    \begin{split}
        {\rm Pr}_D (\widetilde{Y}=y|Y=y, A=a)= &\sum_{c \in \mathcal{C}} {\rm Pr}_D(\widetilde{Y}=y, C=c| Y=y, A=a)\\
        =& \sum_{c \in \mathcal{C}}\frac{{\rm Pr}_D ( \widetilde{Y}=y|Y=y, A=a, C=c) \cdot {\rm Pr}_D(Y=y, A=a, C=c)}{{\rm Pr}_D (Y=y, A=a)}\\
        &= \sum_{c \in \mathcal{C}} \frac{z_{ac}^y \cdot p_{ac}^y}{\alpha_{a}^y}
    \end{split}
\end{equation}
where, $z_{ac}^y= {\rm Pr}_D (\widetilde{Y}=y|Y=y, A=a, C=c)$. As defined in Appendix \ref{app: parameters of LP}$, p_{ac}^y, \alpha^y_a$ are the probability for distribution $D$, $p_{ac}^y = {\rm Pr}_D (Y=y, A=a, C=c), \alpha_a^y= {\rm Pr}_D (Y=y, A=a)$.

The global \textit{Equalized Odds Difference} of predictor $\widetilde{Y}$ for class $y$ is defined as ${\rm Pr}_D(\widetilde{Y}=y|Y=y, A=0)- { \rm Pr}_D (\widetilde{Y}=y|Y=y, A=1)$, which can be expanded as:
\begin{equation}
    \begin{split}
        &{\rm Pr}_D(\widetilde{Y}=y|Y=y, A=0)- { \rm Pr}_D (\widetilde{Y}=y|Y=y, A=1)\\
        =& \sum_{c \in \mathcal{C}}\frac{z_{0c}^y \cdot p_{0c}^y}{\alpha_0^y}-\sum_{c \in \mathcal{C}}\frac{z_{1c}^y \cdot p_{1c}^y}{\alpha_1^y}
    \end{split}
\end{equation}

The first $N$ linear equations of $ - \mathbf{b}\leq\mathbf{A} \mathbf{z} \leq \mathbf{b}$ are: $\forall y \in \mathcal{Y}$,
\begin{equation} \label{eq:30}
    - \epsilon_0 \leq \sum_{c\in \mathcal{C}}\mathbf{n}_{0c}^y - \sum_{c \in \mathcal{C}} \mathbf{n}_{1c}^y \leq \epsilon_0
\end{equation}

Substituting the values of $\mathbf{n}_{0c}^y$ and $\mathbf{n}_{1c}^y$ into Eq. \eqref{eq:30}, Eq. \eqref{eq:30} is:
\begin{equation}
\begin{split}
     -\epsilon_0 \leq \sum_{c \in \mathcal{C}}\frac{z_{0c}^y \cdot p_{0c}^y}{\alpha_0^y} - & \sum_{c \in \mathcal{C}}\frac{z_{1c}^y \cdot p_{1c}^y}{\alpha_1^y} \leq \epsilon_0\\
    \iff - \epsilon_0 \leq  {\rm Pr}_D(\widetilde{Y}=y|Y=y, A=0) - & { \rm Pr}_D (\widetilde{Y}=y|Y=y, A=1) \leq \epsilon_0
\end{split}
\end{equation}

The first $N$ linear equations of $ - \mathbf{b}\leq\mathbf{A} \mathbf{z} \leq \mathbf{b}$ are equivalent to global $\epsilon_0$-global Equalized Odds. Thus, the predictor $\widetilde{Y}$ that satisfies the first $N$ constraints of the convex program also satisfies $\epsilon_0$-global Equalized Odds.

Next, we show that the outcome predictor $\widetilde{Y}$ from Proposition \ref{prop:convex_program} satisfies local Equalized Odds.

The local \textit{Equalized Odds Difference} of predictor $\widetilde{Y}$ for client $c$ and class $y$ is defined as ${\rm Pr}_D (\widetilde{Y}=y |Y=y, A=0, C=c) - { \rm Pr}_D(\widetilde{Y}=y|Y=y, A=1, C=c)$, which is:
\begin{equation}
    \begin{split}
      &{\rm Pr}_D (\widetilde{Y}=y |Y=y, A=0, C=c) - { \rm Pr}_D(\widetilde{Y}=y|Y=y, A=1, C=c) \\  
      = & z_{0c}^y - z_{1c}^y
    \end{split}
\end{equation}

The last $NK$ equations of $ -\mathbf{b}\leq \mathbf{A} \mathbf{z} \leq \mathbf{b}$ are: $\forall y \in \mathcal{Y}, c \in \mathcal{C}$,
\begin{equation}
\begin{split}
    -\epsilon_c \leq z_{0c}^y - & z_{1c}^y \leq \epsilon_c\\
    \iff -\epsilon_c \leq {\rm Pr}_D (\widetilde{Y}=y| Y=y, A=0, C=c) - & {\rm Pr}_D (\widetilde{Y}=y| Y=y, A=1, C=c) \leq \epsilon_c
\end{split}
\end{equation}

The last $NK$ linear equations of $ - \mathbf{b}\leq\mathbf{A} \mathbf{z} \leq \mathbf{b}$ are equivalent to $\epsilon_c$-local Equalized Odds. Thus, the predictor $\widetilde{Y}$ satisfies the last $NK$ equations of the convex program, meaning it satisfies $\epsilon_c$-local Equalized Odds.

The constraint $\mathbf{z}_{ac} \in D_{ac}$ is the condition that any outcome predictor must satisfy. (Proposition \ref{prop:true_positive_feasible}).

The objective function of the convex program is:
\begin{equation*}
    \begin{split}
       &\mathbf{c}^T\mathbf{z}\\
       =& -\sum_{c \in \mathcal{C}} \sum_{a \in \mathcal{A}} \sum_{y \in \mathcal{Y}} z_{ac}^y p_{ac}^y\\
       =& -\sum_{c \in \mathcal{C}} \sum_{a \in \mathcal{A}} \sum_{y \in \mathcal{Y}} {\rm Pr}_D(\widetilde{Y}=y|Y=y, A=a, C=c) {\rm Pr}_D(Y=y, A=a, C=c)\\
        = & -\sum_{y \in \mathcal{Y}} {\rm Pr}_D(\widetilde{Y}=y, Y=y)\\
        = & -{\rm Pr}_D (\widetilde{Y}=Y)
    \end{split}
\end{equation*}
The predictor $\widetilde{Y}$ that minimizes $\mathbf{c}^T\mathbf{z}$ corresponds to maximum accuracy. The maximum accuracy is $-\mathbf{c}^T \mathbf{z}$.

Overall, the predictor $\widetilde{Y}$ satisfies both $\epsilon_0$-global Equalized Odds and $\epsilon_c$-local Equalized Odds while maximizing accuracy. Therefore, the predictor $\widetilde{Y}$ that satisfies the convex program is an optimal $\boldsymbol{\epsilon}$-outcome predictor. The accuracy of the optimal $\boldsymbol{\epsilon}$-outcome predictor is $-\mathbf{c}^T \mathbf{z}$.

\subsection{Proof of Proposition \ref{prop:uniqueness}}
\label{app:proof_uniqueness}
\textbf{Proposition \ref{prop:uniqueness}}: Let $\mathbf{z} \in \mathbb{R}^{2NK}$ be the solution of the LP (\ref{eq: lp}) 
   \begin{eqnarray*}
    \mathbf{z}^T &=& \left[ \begin{array}{ccccccc}
    \mathbf{z}_{01}^T & \mathbf{z}_{11}^T & \mathbf{z}_{02}^T& \mathbf{z}_{12}^T \cdots & \mathbf{z}^T_{0K}& \mathbf{z}^T_{1K} \end{array} \right]
    \end{eqnarray*} and ${\rm TP}_{ac}^y (\widetilde{Y}_{\boldsymbol{1}})$ be the true positive of the predictor  defined in Eq. \eqref{eq:optimal_predictor}.
   For all $a \in \mathcal{A}, c\in \mathcal{C}$, let ${\boldsymbol{\beta}_{ac}}$ be the solution of the following linear algebraic equation (LAE),
   \begin{equation}
       \label{eq:37}
       \mathbf{G}_{ac} \boldsymbol{\beta}_{ac}= \boldsymbol{\gamma}_{ac}
   \end{equation}
   where, the parameter $\mathbf{G}_{ac} \in \mathbb{R}^{(N+1)\times (N+1)}, \boldsymbol{\gamma}_{ac} \in \mathbb{R}^{N+1}$, are detailed in Appendix \ref{app:para_LAE}. Then. the predictor $\widetilde{Y}_{\boldsymbol{\beta}_{ac}}$ that takes value,
         \small
        \begin{equation} \label{eq:38}
    \widetilde{Y}_{\boldsymbol{\beta}_{ac}}(x,a,c)=
        \left\{
      \begin{array}{ll}
      \widetilde{Y}_{\mathbf{1}}(x,a,c), & \text{ with the probability } \beta_{ac}^0 \\[0.2cm]
        y , & \text{with the probability } \beta^y_{ac}, \forall y \in \mathcal{Y}\\[0.2cm]  
      \end{array}
      \right.
    \end{equation}
    \normalsize
    is a fair outcome predictor. There always exists a unique set of parameters $\{\boldsymbol{\beta}_{ac}\}_{\mathcal{A},\mathcal{C}}$, where $\boldsymbol{\beta}_{ac} \in [0,1]^{N+1}$ and $|\boldsymbol{\beta}_{ac}|_{\ell_1}=1$ that satisfies the LAE.
    
\textbf{Proof:} The true positive of class $y$ for  the outcome predictor 
$\widetilde{Y}_{\boldsymbol{\beta}_{ac}}: \mathcal{X} \times \mathcal{A} \times \mathcal{C} \rightarrow \mathcal{Y}$ that takes value of Eq. \eqref{eq:38} is:
\begin{equation}
\begin{split}
    {\rm TP}_{ac}^y(\widetilde{Y}_{\boldsymbol{\beta}_{ac}})
    = {\rm TP}_{ac}^y(\widetilde{Y}_{\mathbf{1}}) \beta_{ac}^0+ \beta_{ac}^y
\end{split}
\end{equation}
Since $\boldsymbol{\beta}_{ac}$ is the solution of LAE \eqref{eq:37}, its elements satisfies that: $\forall y \in \mathcal{Y}$,
\begin{equation}
\begin{split}
    {\rm TP}^{y}_{ac}(\widetilde{Y}_{\mathbf{1}}) \beta_{ac}^{0}+ \beta_{ac}^y &=  z_{ac}^y\\
    \iff {\rm TP}_{ac}^y(\widetilde{Y}_{\boldsymbol{\beta}_{ac}}) &= z_{ac}^y
\end{split}
\end{equation}
The true positives of the predictor $\widetilde{Y}_{\boldsymbol{\beta}{ac}}$ for class $y$, client $c$, and group $a$ are $z_{ac}^y$. Since $\{z_{ac}^y\}_{\mathcal{A}, \mathcal{C}, \mathcal{Y}}$ is the solution of LP \eqref{eq: lp}, it satisfies the fairness constraints of LP \eqref{eq: lp}. Thus, the predictor $\widetilde{Y}_{\boldsymbol{\beta}_{ac}}$ is a fair outcome predictor.

$\mathbf{G}_{ac}$ is a full-rank matrix for all clients and groups, thus, the LAE has a unique solution in all clients and groups.
\subsection{$\widetilde{Y}_{\mathbf{1}}$ maximizes accuracy for distribution D}
\label{app:optimal_of_y_1}
For any outcome predictor $\widetilde{Y}: \mathcal{X} \times \mathcal{A} \times \mathcal{C} \rightarrow \mathcal{Y}$, the accuracy for distribution $D$ is:
\begin{equation}
    \begin{split}
        &{\rm Pr}_D (\widetilde{Y}=Y)\\
        =& \sum_{c \in \mathcal{C}} \sum_{a \in \mathcal{A}}\sum_{y \in \mathcal{Y}} {\rm Pr}_D(\widetilde{Y}=y|Y=y, A=a, C=c) {\rm Pr}_D (Y=y, A=a, C=c)\\
        =& \sum_{c \in \mathcal{C}} \sum_{a \in \mathcal{A}}\sum_{y \in \mathcal{Y}} {\rm TP}_{ac}^y(\widetilde{Y}) {\rm Pr}_D (Y=y, A=a, C=c)\\
        =& \sum_{c \in \mathcal{C}} \sum_{a \in \mathcal{A}} \sum_{y \in \mathcal{Y}} \mathbb{E}_{{\rm Pr}_{X|A,C}}[\mathbf{1}(\widetilde{Y}=y)\cdot r_y(X,a,c)]
    \end{split}
\end{equation}

The predictor that maximizes the accuracy is $\widetilde{Y}_{\mathbf{1}}$ that takes values of:
\begin{equation}
    \widetilde{Y}_{\mathbf{1}}=y, \text{if: } r_y(x,a,c)= \max_{i \in \mathcal{Y}} r_i(x,a,c)
\end{equation}

\section{Experimental Details}
\label{app:exp_details}
\subsection{Data, Models, and Hyperparameters}
We provide the models and hyperparameters of our implementation for each dataset.  
We run the experiments on our local Linux server, equipped with a 16-core 4.00 GHz AMD RYZEN Threadripper Pro 5955WX Processor. All code is implemented in TensorFlow, simulating a federated network with one server and several local clients. 

For the Adult dataset, we randomly divide the data in each local client into three subsets: 60\% for the training set, 20\% for the validation set, and 20\% for the test set. We first implement the \texttt{FedAvg} algorithm. For each communication round in \texttt{FedAvg}, the number of participating communities is set to \(N=2\). The number of local update epochs is set to \(E=1\), with a batch size of \(B=512\). The local models are logistic regression classifiers with two layers, containing 64 and 32 nodes, respectively. We use \textit{ReLU} as the activation function for each hidden layer. These models are trained using the Adam optimizer with a learning rate of \( \eta=0.001 \). We select the number of rounds that minimizes the validation loss . We construct and solve the LP \eqref{eq: lp} and LAE \eqref{eq:ugh-ugh} using the validation data. Finally, we apply post-processing to the test dataset based on the solution from LAE and report its evaluation metrics.

For the PublicCoverage dataset, we similarly split the local dataset into 60\% for training, 20\% for validation, and 20\% for testing. The number of participating communities for \textit{FedAvg} is set to \(N=50\). The number of local update epochs remains at \(E=1\), with a batch size of \(B=256\). We follow the same model structure, optimization algorithm, and evaluation process as with the UCI Adult dataset and report the evaluation metrics.

For the HM 10000 dataset, we similarly split the local dataset into 60\% for training, 20\% for validation, and 20\% for testing. The four diagnostic classes are: class 1 is Pre-cancerous and cancerous lesions, which include 'akiec', 'bcc', and 'mel'; class 2 is Benign lesions, which include 'bkl' and 'df'; class 3 is Nevus-like lesions, which include 'nv'; and class 4 is Vascular lesions, which include 'vasc'. We resize the images to \(28 \times 28 \times 3\). The number of local update epochs is set to \(E=1\), with a batch size of \(B=32\). The local models are CNNs that consist of three convolutional layers (with 32, 64, and 128 filters, respectively), followed by global average pooling and two dense layers (with 128 and 32 nodes, respectively). The loss function used for the multi-class classification task is sparse categorical cross-entropy. The CNN models are trained using the Adam optimizer with a learning rate of \(\eta= 0.0001\). We select the number of rounds that minimizes the validation loss. We construct the LP and LAE using the  validation data, and finally, we apply post-processing to the test dataset based on the solution from the LAE and report its evaluation metrics.

\subsection{Baselines}
\label{app:baselines}
We introduce the baselines used in the experimental section. All baselines are in-processing techniques that modify the optimization algorithms in FL.
\begin{enumerate}
    \item \texttt{FCFL} \citep{cui2021addressing} is designed to achieve local fairness and performance consistency in FL. It achieves fairness by solving a constrained min-max program. We obtained our results by directly running their implementation  at: \href{https://github.com/cuis15/FCFL}{https://github.com/cuis15/FCFL}.
    
    \item \texttt{Fair-Fate} \citep{salazar2023fair} is designed to achieve global fairness. It enforces fairness by adding a momentum term that helps to overcome the oscillations of non-fair gradients. Following the recommendations in \citep{salazar2023fair}, the parameters of the algorithm $\{\lambda_0, \rho, \text{MAX}, \beta_0\}$ are set to $\{0.5, 0.05, 1, 0.99\}$ for the Adult dataset and $\{0.5, 0.05, 1, 0.9\}$ for ACSPublicCoverage.
    
    \item \texttt{FairFed} \citep{ezzeldin2023fairfed} is designed to achieve global fairness. It enforces fairness by adaptively adjusting the aggregation weights of different clients based on their local fairness metrics. The parameter that controls the fairness budget for each update, $\beta$, is set to 1 for both the Adult and ACSPublicCoverage datasets.
    
    \item \texttt{EquiFL} \citep{makhija2024achieving} is designed to achieve both local and global fairness. It enforces fairness by adding a regularization term to the local loss function. The weight for fairness regularization, $w$, is set to $10^2$ for the Adult dataset and $10^5$ for ACSPublicCoverage.
\end{enumerate}

\subsection{Differential Privacy } 
\label{app:pravicy}
The statistics computed and transmitted by the client $c$, as described in Eq.\eqref{eq: est_statistics} of the paper, are:
{\small
\begin{equation} \label{eq:est_statistics}
    \begin{split}
        &{\rm Pr}_D (\widetilde{Y}_{\mathbf{1}} = y, Y = y, A = a \mid C = c) \\ 
        &= \frac{\# \text{ of samples with } (\widetilde{Y}_{\mathbf{1}} = y, Y = y, A = a) \text{ in client } c}{\# \text{ of samples in client } c}, \\
        &{\rm Pr}_D (Y = y, A = a \mid C = c) \\ 
        &= \frac{\# \text{ of samples with } (Y = y, A = a) \text{ in client } c}{\# \text{ of samples in client } c}.
    \end{split}
\end{equation}}

The statistics sent by the client $c$ after applying the Laplace Mechanism are:
\begin{equation}
    \begin{split}
        &{\rm Pr}_D (\widetilde{Y}_{\mathbf{1}} = y, Y = y, A = a \mid C = c) + \mathrm{Lap}(0|b_c), \\
        &{\rm Pr}_D (Y = y, A = a \mid C = c) + \mathrm{Lap}(0|b_c),
    \end{split}
\end{equation}
where $\mathrm{Lap}(x|b) = \frac{1}{2b} e^{-\frac{|x|}{b}}$ is the density function of the Laplace distribution, and $b_c$ is the scale parameter of the Laplace distribution. The larger $b_c$, the greater the variance of the added noise.

The statistics will satisfy $\epsilon$-differential privacy \citep{dwork2006differential},  if we set the scale parameter $b_c$ as:
\begin{equation}
    \begin{split}
        b_c=\frac{\Delta f_c}{\epsilon}
    \end{split}
\end{equation}
where $\Delta f_c$ represents the sensitivity for client $c$, which is the maximum difference in the statistics sent by the client when a single data point is added or removed. 

In our setting, the sensitivity is:
\begin{equation}
    \Delta f_c = \frac{1}{\# \text{ of samples in client } c}.
\end{equation}

Thus, for each client $c$, the scale parameter $b_c$ is given by:
\begin{equation}
    b_c = \frac{1}{(\# \text{ of samples in client } c) \cdot \epsilon}.
\end{equation}

A larger $b_c$ corresponds to a smaller $\epsilon$, providing better privacy protection. To maintain the same level of privacy across all clients, clients with fewer samples will be given a larger $b_c$.

\end{document}